\documentclass[transmag]{IEEEtran}
\usepackage{graphicx}
\usepackage{url}

\usepackage{array}
\usepackage{makecell}
\usepackage{cite}
\usepackage{tabularx}
\usepackage{tablefootnote}
\usepackage{placeins}
\usepackage{amsmath}
\usepackage{hyperref}

\ifCLASSINFOpdf
\else
\fi
\begin{document}
\title{Autonomous Vehicles on the Edge: \\ A Survey on Autonomous Vehicle Racing}

\author{ Johannes Betz, Hongrui Zheng, Alexander Liniger, Ugo Rosolia, Phillip Karle,\\ Madhur Behl, Venkat Krovi, Rahul Mangharam

%\IEEEauthorblockN{Johannes Betz\IEEEauthorrefmark{1},
%Hongrui Zheng\IEEEauthorrefmark{1},
%Alexander Liniger\IEEEauthorrefmark{2},
%Ugo Rosolia\IEEEauthorrefmark{3}, \\
%J. Christian Gerdes\IEEEauthorrefmark{4}, \\
%Panagiotis Tsiotras\IEEEauthorrefmark{4},
%Madhur Behl\IEEEauthorrefmark{5},
%Venkat Krovi\IEEEauthorrefmark{6} and
%Rahul Mangharam\IEEEauthorrefmark{1}} 
%\IEEEauthorblockA{\IEEEauthorrefmark{1} Department of Electrical and Systems Engineering, 
%University of Pennsylvania, Philadelphia, USA}
%\IEEEauthorblockA{\IEEEauthorrefmark{2}Department of Information Technology and Electrical Engineering, ETH Zurich, Switzerland}
%\IEEEauthorblockA{\IEEEauthorrefmark{3} AMBER Lab, California Institute of Technology, Pasadena, USA}
%\IEEEauthorblockA{\IEEEauthorrefmark{4}Department of Mechanical Engineering,
%Stanford University, Stanford, USA}
%\IEEEauthorblockA{\IEEEauthorrefmark{4}School of
%Aerospace Engineering, Georgia Institute of Technology, Atlanta, USA }
%\IEEEauthorblockA{\IEEEauthorrefmark{5}Department of Engineering Systems and Environment, University of Virginia, Charlottesville, USA}
%\IEEEauthorblockA{\IEEEauthorrefmark{6}College of Engineering, Computing and Applied Sciences, Clemson University, Clemson, USA}


\thanks{Manuscript received XX 2021; revised XX XX, 2021. 
Corresponding author: Johannes Betz (email: joebetz@seas.upenn.edu)}
\thanks{This work was sponsored by the  Department of Transportation and NSF}
\thanks{J. Betz, H. Zheng and R. Mangharam are with the Department of Electrical and Systems Engineering, University of Pennsylvania, Philadelphia, USA (e-mail: joebetz, hongruiz, rahulm@seas.upenn.edu) }
\thanks{A. Liniger is with the Department of Information Technology and Electrical Engineering, ETH Zurich, Switzerland (e-mail: alex.liniger@vision.ee.ethz.ch)}
\thanks{U. Rosolia is with the AMBER Lab, California Institute of Technology, Pasadena, USA (e-mail: urosolia@caltech.edu)}
\thanks{P. Karle is with the Institute of Automotive Technology, Technical University of Munich, Germany (e-mail: phillip.karle@tum.de)}
\thanks{M. Behl is with the Department of Computer Science, University of Virginia, Charlottesville, USA (e-mail:madhur.behl@virginia.edu)}
\thanks{V. Krovi is with the College of Engineering, Computing and Applied Sciences, Clemson University, Clemson, USA (e-mail:mb2kg@virginia.edu)}
}

\markboth{$>$ REPLACE THIS LINE WITH YOUR PAPER IDENTIFICATION NUMBER $<$}
{$>$ REPLACE THIS LINE WITH YOUR PAPER IDENTIFICATION NUMBER $<$}



% for Transactions on Magnetics papers, we must declare the abstract and
% index terms PRIOR to the title within the \IEEEtitleabstractindextext
% IEEEtran command as these need to go into the title area created by
\IEEEtitleabstractindextext{%
\begin{abstract}
The rising popularity of self-driving cars has led to the emergence of a new research field in the recent years: Autonomous racing. Researchers are developing software and hardware for high performance race vehicles which aim to operate autonomously on the edge of the vehicles limits: High speeds, high accelerations, low reaction times, highly uncertain, dynamic and adversarial environments. This paper represents the first holistic survey that covers the research in the field of autonomous racing. We focus on the field of autonomous racecars only and display the algorithms, methods and approaches that are used in the fields of perception, planning and control as well as end-to-end learning. Further, with an increasing number of autonomous racing competitions, researchers now have access to a range of high performance platforms to test and evaluate their autonomy algorithms. This survey presents a comprehensive overview of the current autonomous racing platforms emphasizing both the software-hardware co-evolution to the current stage. Finally, based on additional discussion with leading researchers in the field we conclude with a summary of open research challenges that will guide future researchers in this field.
\end{abstract}

\begin{IEEEkeywords}
Autonomous systems,  autonomous vehicles, intelligent vehicles, advanced driver assistance, simultaneous localization and
mapping, path planning, control
\end{IEEEkeywords}}

\maketitle

\IEEEdisplaynontitleabstractindextext

% \fi
\IEEEpeerreviewmaketitle

\section{Introduction}
\IEEEPARstart{W}{hat} aerospace engineering is to aviation, motorsport is to automotive technology. For over a century now, racing series such as Formula 1, Indy Car or the World Rally Championship have served to inspire research and product innovation to improve performance and safety in commercial road vehicles. These developments include well-known elements such as the disc brake, the turbocharger or production measures for fibre composites (e.g. carbon). In more recent years, developments in the hybrid powertrain but also in the connectivity (real-time measurement and transfer of vehicle data) of the vehicles have emerged. With millions of dollars of investment and prestige at stake, these developments target towards a singular goal: The racecar must achieve the fastest lap time and thus win the race at the end. With a vehicle that moves at the dynamic limits of handling, that reaches high velocities, that is designed for aerodynamic efficiency and that consists of pure lightweight construction, traditionally the target of the minimum lap time can only be achieved through extremely novel, sophisticated and radical developments. But the technical development of the racecar is only half of the effort. 
In motorsport racing, it all boils down to the ability of the driver to operate the racecar at its limits  \cite{Doubek2021,Lima2020}. Expert race drivers are extremely proficient in pushing the racecar to its dynamical limits of handling, while account for continuously changing parameters such as tire wear, changing brake bias, and engine maps from turn to turn, communicating strategy and status with the team in the pits, and trying to maintain track position or overtake fierce competitors all while driving at speeds exceeding 300 km/h. \\
\indent All of these facets manifest themselves in the innumerable research and development challenges sought to be addressed in the burgeoning field of \textit{autonomous vehicle racing}. The current state of the art of autonomous driving software -- either from commercial companies or researchers -- is capable of operating autonomously but only to a limited velocity. Everything that we find in classic motorsport can also be found in autonomous motorsport - with one difference: The racecar-driver is based on software only. More or less this means that a highly sophisticated autonomous driving software needs to replace the human pilot and it should be capable of detecting other vehicles, localize the vehicle position relative to the opponents and the track while driving at high speeds,  planning dynamic trajectories to allow overtaking in adversarial environments, and correcting at high frequency to the steering angle to stay on the racetrack. Furthermore, the vehicle needs to execute a performance assessment on its own by adjusting the aerodynamics, energy distribution, differential settings or brake balance settings based on tire wear, temperature and weather. On this premise, an autonomous racecar exceeds the requirements for the software to a wide extent in comparison to a normal passenger car - which provides many learning outcomes, research questions and new algorithmic developments \cite{Betz2018,Jiang2021}. It is also worthy of note that the requirements for autonomous racecars versus passenger vehicles can be quite disparate - precluding a direct transplant of a complete autonomous driving stack. Therefore, autonomous racing has emerged as a field where advanced algorithmic approaches are tested and then individually transferred to the development of autonomous passenger vehicles - similar to classic motorsport. \\
\indent This paper provides the first survey of the state of the art research in the field of autonomous vehicle racing. By summarizing, classifying and evaluating the different software and hardware developments we provide a holistic overview of the research in this field. Finally, we discuss future research directions by highlighting open questions and challenges in autonomous racing.

\nocite{*}

\subsection{Contributions}
In this survey we present the efforts and the research that was conducted in recent years in the field of autonomous racing. This work has four main contributions:
\begin{enumerate}
    \item With this paper we provide the first survey to comprehensively cover the topic of autonomous vehicle racing for both software and hardware developments.
    \item We provide an extensive review of all research papers that developed new autonomy software for autonomous racecars. By splitting this software review into subsections of perception, planning and control we display in detail which methods and approaches were used. We compare the different approaches to each other and explain their algorithmic setup. Furthermore, we discuss recent efforts made with methods from the field of deep neural networks (DNN) and reinforcement learning (RL) to achieve a partial or full end-to-end pipelines for autonomous racing.
    \item We display an overview of the current autonomous racing competitions which provide hardware, a racing environment as well as an organization (e.g. sports and technical regulations).  We compare the different racing series and hardware against each other and give a holistic overview for potential interested researchers.
    \item Finally, we present a list of open research questions and challenges in the field of autonomous racing. We discuss that these open challenges can be applied to the field of passenger cars, too, and provide opportunities for future researchers to work on relevant research topics.
\end{enumerate}
\vspace{-0.3cm}
\subsection{Preliminary Remarks}
\subsubsection{Definition: Autonomous Racing}
Although the term of autonomous racing can be referred to different applications (e.g. drone racing) we focus in this paper only on research in the field of autonomous racing \emph{cars}. These racecars need to have four wheels, can either have a combustion engine or electrical engine as a main power unit, can be real racecars (e.g. Formula 1 car) or small-scale vehicles (e.g. 1:10 scale). In addition, the software and hardware surveyed here must have a clear connection to the field of automobile racing. This means that the authors of these papers either used a specific hardware that is acting in a racing environment (e.g. racetrack, adversarial setup), they used a specific simulation that displays a racecar within an racing environment (e.g. racing game for PC) or their research displays specific solutions for a racing problem (e.g. driving fast autonomously around the racetrack). Although some authors present results and algorithms for high speed autonomous driving on the freeway this work is not covered in this survey because neither the aspect of handling the vehicles at the limits nor the adversarial context is given here sufficiently.
\subsubsection{Research Categorization}
\begin{figure}[h]
\begin{center}
\includegraphics[scale=0.95]{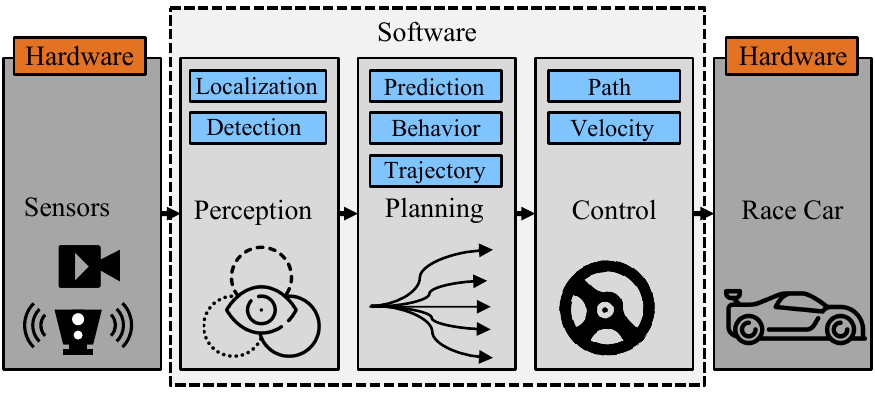}
\caption{Autonomous driving pipeline including both hardware and software that provides the categorization for the survey in the field of autonomous racing}
\label{fig_pipeline}
\end{center}
\end{figure}

The field of autonomous racing provides plenty of development and research categories. For this survey we want to display both software and hardware efforts in the field of autonomous racing and therefore we use the following  \textit{perception - planning - control} pipeline \cite{pendlton2017} depicted in Figure  \ref{fig_pipeline} to categorize the research papers.

\vspace{-0.3cm}
Research and developments in the field of autonomous racing hardware (sensors and vehicle hardware) are discussed in Section \ref{sec_hardware}. Unfortunately, we have not seen any specific developments of sensors for the purpose of autonomous racing and therefore we heavily focus on the used racing vehicle platforms. By presenting vehicle setups (sensors, computation hardware, racing environment) we provide a clear overview of which hardware is available for researchers. The biggest part of this survey paper presents research and software in the field of perception, planning and control in Section \ref{sec_software}. In Subsection \ref{subsec_perception} \textit{perception}, we cover all algorithms that provide either a solution for mapping, localization or object detection. In Subsection \ref{subsec_planning}  \textit{planning} we display global and local trajectory and behavior planners. The final Subsection \ref{subsec_control} is used to present algorithms in the field of \textit{control} and displays the solutions for path and velocity tracking at the handling limits. Unfortunately, there are many papers out there that act on the intersection of planning and control. Those papers have no clear distinction in which field they belong and therefore we decided to categorize them in either one of those field based on their focus. The method of DNNs has become more and more popular in the recent years and different authors proposed so called \textit{end-to-end} approaches that solve the autonomous driving task. These kind of techniques are displayed in Subsection \ref{subsec_end2end}. In addition, some authors proposed evaluations with racecars,  complete software pipelines, modelling efforts and simulation environments for autonomous racecars that do not fit in the proposed categories. Those papers are listed in Subsection \ref{subsec_additioalsoftware}. In summary, this survey is covering all research papers in the field published until the end of 2021.

\section{Autonomous Racing Software}
\label{sec_software}

\subsection{Perception}
\label{subsec_perception}

Perception provides the general term for all algorithms that perceive the environment and derive knowledge about it. In particular, perception includes detecting objects, detecting the free space, mapping the environment as well as localizing the autonomous vehicle. In an autonomous racing environment we deal with high speeds and therefore the question arises: How fast is too fast? Falanga et al.~\cite{Falanga2019} tried to answer this question for autonomous robots with an additional case study on autonomous quadrocopters. The authors came to the conclusion that the maximum latency an autonomous system can tolerate to guarantee safety (not crashing in an object) is related to the desired speed, the agility of the system (e.g. the maximum acceleration it can produce) and the perception parameter of the sensors (e.g. the the sensing range). For autonomous racecars the same parameters can be taken into account but no particular evaluation regarding high speed perception for autonomous racecars has been done yet. The current state of the art in the field of autonomous racing perception is summarized, categorized and displayed in Table~\ref{tab:software_perception}. 

\begin{table*}[h]
\centering
\caption{Overview of research in the field of autonomous racing perception}
\label{tab:software_perception}

\begin{tabular}{|>{\raggedleft\arraybackslash}m{2.9cm}|>{\centering\arraybackslash}m{0.5cm}|>{\centering\arraybackslash}m{2.0cm}|>{\centering\arraybackslash}m{2.5cm}|>{\centering\arraybackslash}m{2.2cm}|>{\centering\arraybackslash}m{1.2cm}|>{\centering\arraybackslash}m{1.3cm}|>{\centering\arraybackslash}m{1.3cm}|}
\hline
\textbf{Name and Reference} & \textbf{Year} & \textbf{\begin{tabular}[c]{@{}c@{}}Perception \\ Category\end{tabular}} & \textbf{Method} & \textbf{\begin{tabular}[c]{@{}c@{}}Sensor \\ Type\end{tabular}} & \textbf{\begin{tabular}[c]{@{}c@{}}Tested on\\ Hardware\end{tabular}} & \textbf{\begin{tabular}[c]{@{}c@{}}Racing \\ Series\end{tabular}} & \textbf{\begin{tabular}[c]{@{}c@{}}Max. Speed \\ (km/h)\end{tabular}} \\ \hline
Nobis et al.~\cite{Nobis2019}               & 2019  & Mapping               & SLAM & LiDAR & Yes & Roborace & 30   \\ \hline
Palafox et al.~\cite{Palafox2019}           & 2019  & Free Space Detection  &  \begin{tabular}[c]{@{}c@{}}Semantic \\ Segmentation \end{tabular} & Camera & No & Roborace & -   \\ \hline
Valls et al.~\cite{Valls2018}     & 2018  & Localization          & SLAM, EKF & LiDAR; Odometry  & Yes & FSD & 80   \\ \hline
Brunnbauer et al.~\cite{Brunnbauer2019}     & 2019  & Localization          & Cone Detection & Camera  & Yes & F1TENTH & -   \\ \hline
Gotlieb et al.~\cite{Gotlib2019}            & 2019  & Localization          & AMCL & LiDAR; Odometry & Yes & F1TENTH & -   \\ \hline
Stahl et al.~\cite{Stahl2019}               & 2019  & Localization          & AMCL  & LiDAR & Yes & Roborace & 150   \\ \hline
Wischnewski et al.~\cite{Wischnewski2019}   & 2019  & Localization          & Kalman Filter & GPS; LiDAR; Odometry & Yes & Roborace & 150   \\ \hline
Massa et al.~\cite{Massa2020}               & 2020  & Localization          & EKF, AMCL & LiDAR; Odometry & Yes & Roborace & 200   \\ \hline
Renzler et al.~\cite{Renzler2020}           & 2020  & Localization          & Distortion correction  & LiDAR & Yes & Roborace & 90   \\ \hline
Zubaca et al.~\cite{Zubaca2020}             & 2020  & Localization          & H $\infty$ Filter & LiDAR; Odometry & Yes & Roborace & 60   \\ \hline
Schratter et al.~\cite{ICRAworkshop_04}     & 2021  & Localization          & NDT  & LiDAR & Yes & Roborace & 122   \\ \hline 
Gosala et al.~\cite{Gosala2019}             & 2019  & Localization          & SLAM & LiDAR; Odometry  & Yes & FSD & 90    \\ \hline
Andresen et al.~\cite{Andresen2020}         & 2020  & Localization          & GraphSLAM; Cone Detection  & Camera; LiDAR; Odometry  & Yes & FSD & -  \\ \hline
Srinivasan et al.~\cite{Srinivasan2020}     & 2020  & Localization          & RNN & Odometry; Vehicle Data & Yes & FSD & 40   \\ \hline
Le Large et al.~\cite{LeLarge2021}          & 2021  & Localization          & Graph SLAM; EKF SLAM & LiDAR  & Yes & FSD & -   \\ \hline
Peng et al.~\cite{Peng2021}                 & 2021  & Localization          & GraphSLAM; Cone Detection & Camera; LiDAR; Odometry  & Yes & FSD & -  \\ \hline
%Sauerbeck et al.~\cite{Sauerbeck2022}       & 2022  & Localization          & SLAM, Camera & Camera,LiDAR & Yes & IAC & 220 \\ \hline
Strobel et al.~\cite{Strobel2020}           & 2020  & Object Detection \& Localization & YOLO v3 & Stereo camera and Monocular camera & Yes & FSD & -  \\ \hline
Dhall et al.~\cite{Dhall2019}               & 2019  & Object Detection      & YOLO v2 & Camera & Yes & FSD & -  \\ \hline
De Rita et al.~\cite{DeRita2019}            & 2019  & Object Detection      & Tiny-YOLO; Proteins & Camera & Yes & FSD & - \\ \hline
Puchtler et al.~\cite{Puchtler2020}         & 2020  & Object Detection      & SSD MobileNet v2 & Camera & Yes & FSD & -   \\ \hline
Dodel et al.~\cite{dodel2021}               & 2021  & Object Detection      & Dataset & Camera & Yes & FSD & - \\ \hline
\end{tabular}
\end{table*}

\begin{figure}[h]
\begin{center}
\includegraphics[width=8cm]{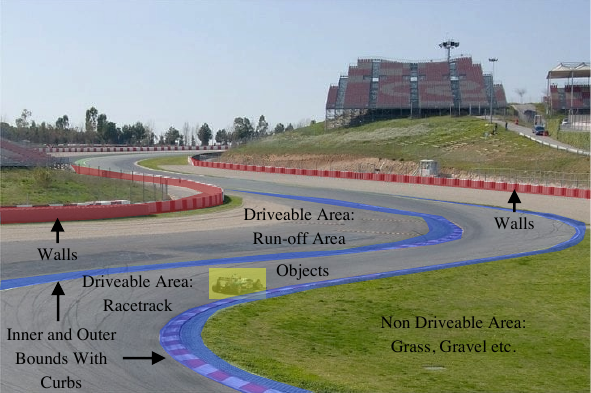}
\caption{Racetrack with environmental specifics: Inner-and outer bounds, racecar objects, walls and run-off area}
\label{fig_racetrack}
\end{center}
\end{figure}

A racetrack normally provides some specifics that can not be found on normal streets. As depicted in Figure  \ref{fig_racetrack}, a racetrack consists of a single lane that is the main driveable space. This lane is defined by an inner and outer bound that can consists of additional curbs in the turns. On both left and right side of the track there are zones consisting of grass, gravel or tarmac (run-off area) where racecars can drive in an evasive maneuver or if they miss the race line. The racetrack is finally surrounded by walls. Depending on the racing series these types of the racetrack features can vary (e.g. Formula E: no gravel or grass). We define the fundamental problems for autonomous racing perception as the following:
\begin{itemize}
    \item High speed object detection.
    \item High speed localization and state estimation.
    \item Localization on wide areas without specific landmarks.
    \item Precise localization information necessary to achieve high dynamic trajectory planning and control.
\end{itemize}

\begin{figure}[h!]
\begin{center}
\includegraphics[width=8.0cm]{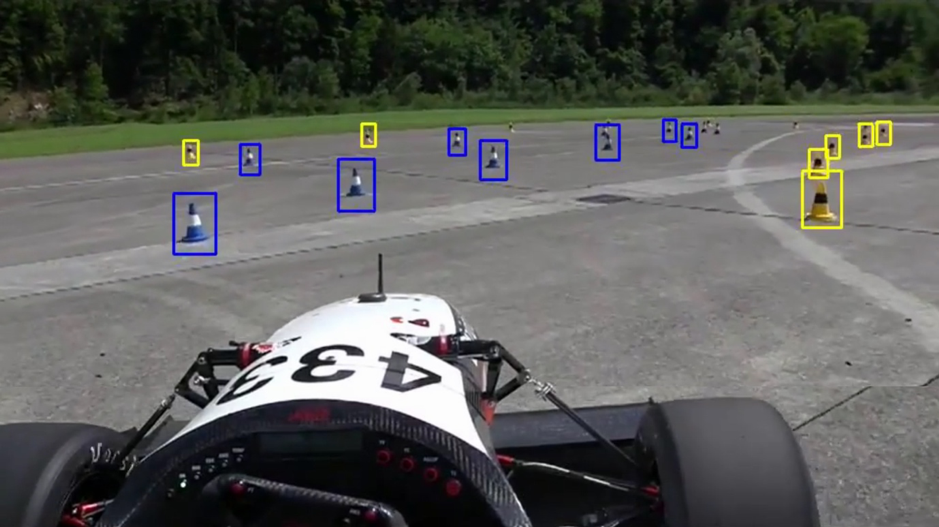}
\caption{Exemplary cone detection in the Formula Student Driverless competition. The racetrack is defined by a left boundary (blue cones) and a right boundary (yellow cones) which need to be detected by the teams.}
\label{fig_perception_FSD_cones}
\end{center}
\end{figure}

Although the racetrack provides a very simple structure with a single driveable lane, the long distance to the walls and non-existing landmarks make this environment quite difficult to perceive. None of the displayed papers are using precreated High-Defintion Maps (HD-Maps) that are known from passenger autonomous driving development.  An open-source library \cite{Heilmeier2019_2} of racetracks provides a simple 2D-birds-eye-view with inner and outer bounds ($x$-and $y$-Position) of about 30 racetracks around the world that can be used for planning but not for localization. In Nobis et al.~\cite{Nobis2019} an adaption and enhancement of well-known simultaneous localization and mapping (SLAM) algorithms (\textit{Google Cartographer}\cite{Cartographer2016}, \textit{GMapping}) are displayed to create a map of large-scale outdoor environments. Palafox et al.~\cite{Palafox2019} use a vision-based method to detect the free space when no lane lines are present by only using camera images and depth information as input for a DNN. \\
\indent Most perception research for autonomous racing is listed in the field of localization techniques. Although many research vehicles are equipped with differential GPS (dGPS) that delivers a high localization accuracy, the goal of many autonomous racing researchers is to deliver software based localization solutions only. Both \cite{Brunnbauer2019} and \cite{Gotlib2019} are using a 1:10 scale vehicle for their localization techniques. While Brunnbauer et al.~\cite{Brunnbauer2019} use the camera to detect cones that create features to enhance the odometry localization, Gotlieb et al.~\cite{Gotlib2019} map the track with an onboard 2D-LiDAR and run a Robot Operating System (ROS) based Adaptive Monte Carlo Localization (AMCL). The same AMCL approach is also used and adapted by Stahl et al.~\cite{Stahl2019} to run on the Roborace research vehicle. By using pregenerated maps based on LiDAR data, the car achieves a mean absolute lateral error of 0.086m at a velocity of 150 km/h. A comparison to this work is done in \cite{Wischnewski2019} were odometry, GPS and LiDAR data is fused in a Kalman Filter (KF) based on a purely kinematic vehicle  dynamics model to achieve localization at high speeds. The Roborace vehicle is also used for the localization research of Renzler et al.~\cite{Renzler2020}, Zubaca et al.~\cite{Zubaca2020} and Schratter et al.~\cite{ICRAworkshop_04}. To increase the localization performance at high speeds the distortion of the LiDAR measurement is analyzed and a compensation is proposed in \cite{Renzler2020}. It is shown that this correction can be implemented straightforward and has a high benefit for objects moving at faster speeds. The work from Zubaca et al.~\cite{Zubaca2020}  presents an extended H$\infty$ Filter (EHF) based on a kinematic motion model assuming constant turn-rate and acceleration to fuse LiDAR, IMU (inertial measurement unit), and vehicle dynamic sensors' measurements. The proposed EHF shows slightly better estimation performance in high dynamic driving scenarios in comparison to an extended Kalman Filter (EKF). In \cite{ICRAworkshop_04} a complete process for both mapping an localization on racetracks with the Normal Distributions Transform (NDT) method is displayed. Based on this approach the Roborace vehicle reaches up to 122 km/h and an average localization error of 0.2~m. Massa et al.~\cite{Massa2020} are using two multi-rate EKFs and an extend AMCL that exploits some a priori knowledge of the environment on the Roborace vehicle.
\newpage
\noindent The authors showed that the pose error heavily depends on the car's velocity, and varies in average from 0.1~m (at 60~km/h) to 1.48~m (at 200~km/h) laterally and from 1.9~m (at 100~km/h) to 4.92~m (at 200~km/h) longitudinally. \\
\indent For Formula Student Driverless (FSD) vehicles Le Large et al.~\cite{LeLarge2021} show a comparison between GraphSLAM and an EKF-SLAM. Based on their experimental analysis with the FSD vehicle and the maps generated by the algorithms they showed that GraphSLAM outperforms EKF-SLAM in terms of accuracy. In \cite{Valls2018,Gosala2019,Andresen2020} the localization and mapping approaches for the FSD vehicle of the AMZ team is presented. While in \cite{Gosala2019} only a LiDAR based SLAM was used, the student team extended the work with an additional LiDAR and camera based object detection for cones on the track \cite{Andresen2020}. With this setup the team achieved a velocity of 10~m/s while doing mapping, localization and planning at the same time with an RSME error of 0.29~m. On the same vehicle a recurrent neural network (RNN) was applied \cite{Srinivasan2020} to derive an accurate velocity estimation. By taking different vehicle sensors (e.g. IMU, wheel encoder) into account this learning based approach reached 15x better performance than an EKF approach with an RSME of $v_x$~=~0.141~m/s and $v_y$~=~0.059~m/s. \\
\indent The current state of the art in autonomous racing is heavily based on single vehicle races. Therefore the subcategory of object detection algorithms for high speed applications was not given much attention. Nevertheless, in the FSD competition teams need to detect both color and form of cones to let the vehicle drive autonomously as depicted in Figure  \ref{fig_perception_FSD_cones}. In \cite{DeRita2019} a case study with different convolutional neural network (CNN) methods (Tiny-YOLO, Proteins) are done in comparison to a YOLO v2 setup \cite{Dhall2019} to display the best approach for cone detection in the FSD scenario. Strobel et al.~\cite{Strobel2020} present a combination of a YOLO v3 based object detection, pose estimation, and time synchronization that uses data from both stereo and monocular cameras. Furthermore, besides these software focused developments the authors of \cite{Puchtler2020} evaluated the performance and energy consumption of popular, off-the-shelf commercial devices for DNN inference in the formula student context. Finally, to help  and support other FSD teams in their development,  Dodel et al.~\cite{dodel2021} presented a collaborative dataset for vision-based cone detection systems that is open-source available.

\subsection{Planning}
\label{subsec_planning}

In the following subsection we cover algorithms that plan trajectories for the autonomous race vehicle to drive around the racetrack. Strategies that drive the vehicle end-to-end directly from perception to actuation is excluded from this part and is discussed with further details in Subsection~\ref{subsec_end2end}. We split the discussion into the three following parts. 

\begin{figure}[h]
\begin{center}
\includegraphics[scale=1.25]{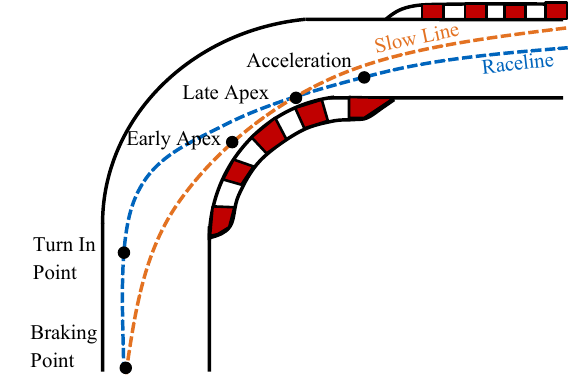}
\caption{Theory of raceline (blue) and slow line (orange) on the racetrack.The raceline provides the global fastest path around the complete racetrack.}
\label{fig_planning_raceline}
\end{center}
\end{figure}

\textit{Global planning} provides an optimal path, better known as \textit{raceline} (depicted in Figure  \ref{fig_planning_raceline}), around the racetrack. In the context of racing, global planning often optimize for the lowest lap time. Therefore, when following this raceline, the car drives an optimal path around the racetrack -- under the constraints of the raceline generation -- as fast as possible. 

\indent \textit{Local planning} (or \textit{motion planning}) plans on a finer granularity compared to global planning, usually under the assumption that an optimal global trajectory is provided. Local planners operate in a certain time horizon, and aim to avoid obstacles while still provide a fast and reliable path that does not deviate too much from the optimal global raceline.
\indent Finally, \textit{behavioral planning} provides information about the high-level mission planning of the racecar. This can include the decision making about overtaking maneuvers (overtaking left/overtaking right/stay behind), the energy management strategy, interaction with other vehicles and the reaction to inputs from race control (e.g. flags, speed limits). As a summary, Table \ref{tab:software_planning} provides an overview of research efforts in the field of planning for autonomous racing. We define the fundamental problems for autonomous racing planning as the following:
\begin{itemize}
    \item Minimum-time optimization for a global optimal raceline. 
    \item Long local planning horizon for recursive feasibility.
    \item Obstacle avoidance and vehicle reaction at high speeds.
    \item High re-planning frequency for real-time capability.
    \item Decision making under high uncertainty.
    \item Interaction planning with non-cooperative agents.
\end{itemize}

\begin{table*}[h]
\centering
\caption{Overview of research in the field of autonomous racing planning}
\label{tab:software_planning}

% \begin{tabular}{|>{\raggedleft\arraybackslash}m{3.0cm}|>{\centering\arraybackslash}m{0.5cm}|>{\centering\arraybackslash}m{3.0cm}|>{\centering\arraybackslash}m{5.0cm}|>{\centering\arraybackslash}m{1.5cm}|>{\centering\arraybackslash}m{1.1cm}|>{\centering\arraybackslash}m{2.1cm}|}
\begin{tabular}{|r|c|c|c|c|c|}
\hline
% \textbf{Name and Reference} & \textbf{Year} & \textbf{\begin{tabular}[c]{@{}c@{}}Planning \\ Category\end{tabular}} & \textbf{Method} & \textbf{\begin{tabular}[c]{@{}c@{}}Tested on\\ Hardware\end{tabular}} & \textbf{Racing Series} \\ \hline
\textbf{Name and Reference} & \textbf{Year} & \textbf{Planning Category} & \textbf{Method} & \textbf{\begin{tabular}[c]{@{}c@{}}Tested on\\ Hardware\end{tabular}} & \textbf{Racing Series} \\ \hline
Metz et al.~\cite{Metz1989}                      & 1989 & Global Planning         & Near time optimal control            & No  & - \\ \hline
Graghin et al.~\cite{Braghin2008}                & 2008 & Global Planning         & Optimization (Min. Curvature)        & No  & - \\ \hline
Kelly et al.~\cite{Kelly2010}                    & 2010 & Global Planning         & Time optimal control                 & No  & - \\ \hline
Cardamone et al.~\cite{Cardamone2010}            & 2010 & Global Planning         & Optimization (Min. Curvature)        & No  & - \\ \hline
Quadflieg et al.~\cite{Quadflieg2011}            & 2011 & Global Planning         & CMA-ES (Min. time)                   & No  & - \\ \hline
Theodosis et al.~\cite{Theodosis2011}            & 2011 & Global Planning         & Multi-phase Geometric based          & Yes & - \\ \hline
Theodosis et al.~\cite{Theodosis2012}            & 2012 & Global Planning         & Non-linear Optimization (Min. time)  & Yes & - \\ \hline
Rucco et al.~\cite{Rucco2015}                    & 2015 & Global Planning         & Optimization (Min. time)             & No  & - \\ \hline
Bevilacqua et al.~\cite{Bevilacqua2017}          & 2017 & Global Planning         & Particle Swarm Optimization          & No  & - \\ \hline
Kuhn~\cite{Kuhn2017}                             & 2017 & Global Planning         & Geometric methods                    & No  & - \\ \hline
Dal Bianco et al.~\cite{DalBianco2018}           & 2018 & Global Planning         & Optimal Control (Min. time)          & No  & - \\ \hline
Heilmeier et al.~\cite{Heilmeier2019}            & 2019 & Global Planning         & Optimization (Min. Curvature)        & Yes & Roborace \\ \hline
% Gundlach et al. ~\cite{Gundlach2019}             & 2019 & Global Planning         & Optimization (Min. time)             & No  & - \\ \hline
Herrmann et al.~\cite{Herrmann2019,Herrmann2020} & 2019 & Global Planning         & Optimization (Min time+energy)       & Yes & Roborace \\ \hline
Herrmann et al.~\cite{Herrmann2020_2}            & 2020 & Global Planning         & Optimization (velocity)              & Yes & Roborace \\ \hline
Pagot et al.~\cite{Pagot2020}                    & 2020 & Global Planning         & NMPC (Min. time)                     & Yes & - \\ \hline
Vazquez et al.~\cite{Vazquez2020}                & 2021 & Global Planning         & Optimization (Min. time) + NMPC      & Yes & FSD \\ \hline
Lovato et al.~\cite{Lovato2021}                  & 2021 & Global Planning         & Apex-finding + Optimal control       & No  & - \\ \hline
Butz et al.~\cite{Butz2009}                      & 2009 & Local Planning          & CMA                                  & No  & - \\ \hline
Jeong et al.~\cite{Jeong2013}                    & 2013 & Local Planning          & RRT*                                 & No  & - \\ \hline
Liniger et al.~\cite{Liniger2014}                & 2014 & Local Planning          & NMPC                                 & Yes & 1:43 \\ \hline
Liniger et al.~\cite{Liniger2015}                & 2015 & Local Planning          & Viability theory + MPC               & Yes & 1:43 \\ \hline
Anderson et al.~\cite{Anderson2016}              & 2016 & Local Planning          & MPC                                  & No  & - \\ \hline
Kapania et al.~\cite{Kapania2016}                & 2016 & Local Planning          & Convex Optimization                  & Yes & - \\ \hline
Williams et al.~\cite{Williams2016}              & 2016 & Local Planning          & MPPI                                 & Yes & Autorally \\ \hline
Buyal et al.~\cite{Buyval2017}                   & 2017 & Local Planning          & Nonlinear MPC (NMPC)                 & Yes & Roborace \\ \hline
Funke et al.~\cite{Funke2017}                    & 2017 & Local Planning          & MPC                                  & Yes & - \\ \hline
Arslan et al.~\cite{Arslan2017}                  & 2017 & Local Planning          & CL-RRT$^\#$                          & No  & - \\ \hline
You et al.~\cite{You2018}                        & 2018 & Local Planning          & Trail braking                        & No  & - \\ \hline
Subosits et al.~\cite{Subosits2019}              & 2019 & Local Planning          & Convex Optimization                  & Yes & - \\ \hline
Stahl et al.~\cite{Stahl2019_2}                  & 2019 & Local Planning          & Graph search + spline opt.           & Yes & Roborace \\ \hline
Alcal et al.~\cite{Alcal2020_2}                  & 2020 & Local Planning          & LPV-MPC                              & No  & - \\ \hline
Bulsara et al.~\cite{Bulsara2020}                & 2020 & Local Planning          & RRT+MPC                              & Yes & F1TENTH \\ \hline
Feraco et al.~\cite{Feraco2020}                  & 2020 & Local Planning          & RRT + Stanley                        & Yes & FSD \\ \hline
Srinivasan et al.~\cite{srinivasan2021}          & 2021 & Local Planning          & Holistically designed hierarchical controllers                  & Yes & FSD \\ \hline
Evans et al.~\cite{evans2021, evans2021_2}       & 2021 & Local Planning          & TD3                                  & No  & F1TENTH \\ \hline
Kalaria et al.~\cite{ICRAworkshop_08}            & 2021 & Local Planning          & NMPC                                 & No  & - \\ \hline
Brüdigam et al.~\cite{ICRAworkshop_11}           & 2021 & Local Planning          & MPC + Gaussian                       & No  & - \\ \hline
Bhargav et al.~\cite{ICRAworkshop_09}            & 2021 & Local Planning          & MPC                                  & No  & - \\ \hline
You et al.~\cite{You2021}                        & 2021 & Local Planning          & Trail braking                        & No  & - \\ \hline
Wang et al.~\cite{ICRAworkshop_12}               & 2021 & Local Planning          & MPC + Deep-Koopman                   & Yes & F1TENTH \\ \hline
Jung et al.~\cite{ICRAworkshop_01}               & 2021 & Local Planning          & MPC + Game Theory                    & No  & IAC       \\ \hline
Loiacona et al.~\cite{Loiacono2010}              & 2010 & Behavioral Planning     & Reinforcement Learning               & No  & - \\ \hline
%Herrmann et al.~\cite{Herrmann2022}              & 2022 & Behavioral Planning     & Optimization                         & Yes  & Roborace \\ \hline
Kloeser et al.~\cite{Kloeser2020}                & 2020 & Global+Local Planning   & NMPC                                 & No  & - \\ \hline
O'Kelly et al.~\cite{OKelly2020}                 & 2020 & Global+Local Planning   & CMA-ES + Spline opt.                 & Yes & F1TENTH \\ \hline
Williams et al.~\cite{Williams2017}              & 2017 & Local+Behavior Planning & Best response + MPPI                 & Yes & Autorally \\ \hline
Notomista et al.~\cite{Notomista2020}            & 2020 & Local+Behavior Planning & Iterated Best Response + CBF         & No  & - \\ \hline
Wang et al.~\cite{wang2020}                      & 2020 & Local+Behavior Planning & Iterated Best Response               & No  & - \\ \hline
Sinha et al.~\cite{sinha2020}                    & 2020 & Local+Behavior Planning & EXP3 + Dist. robust opt.             & Yes & F1TENTH \\ \hline
Liniger et al.~\cite{Liniger2020}                & 2020 & Local+Behavior Planning & Non-cooperative game                 & Yes & 1:43 \\ \hline
Wang et al.~\cite{Wang2019, Wang2021}            & 2021 & Local+Behavior Planning & Iterated Best Response + Spline opt. & Yes & - \\ \hline
Schwarting et al.~\cite{Schwarting2021_2}        & 2021 & Local+Behavior Planning & Belief-space Planning                & No  & - \\ \hline
\end{tabular}
\end{table*}

\begin{figure}[h!]
\begin{center}
\includegraphics[scale=0.95]{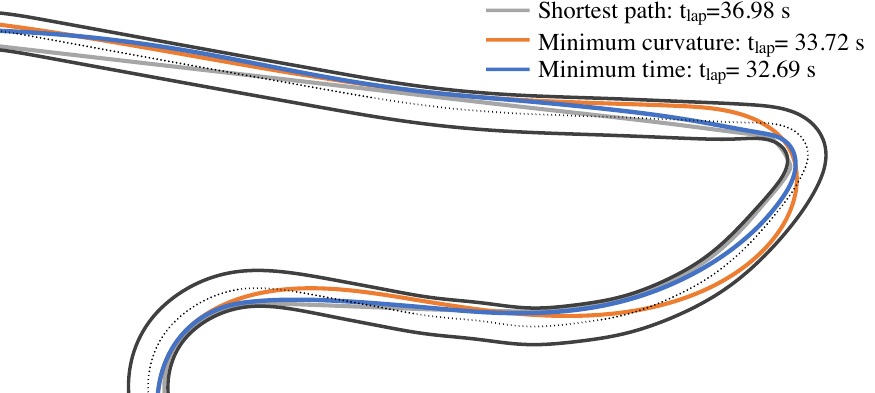}
\caption{Comparison of global optimal raceline algorithms based on shortest path, minimum-curvature \cite{Heilmeier2019} and minimum-time \cite{Christ2019} optimization, which lead to different lap times $t_{lap}$ and trajectories.}
\label{fig_global_racelin_comp}
\end{center}
\end{figure}

% Using the center line works well for narrow tracks, as for example in FSD or in small scale experiments where the planning can be performed online by advanced predictive controllers. However, such approaches fail in full scale, high speed environments, where extremely long prediction horizons in the range of several hundred meter to more would be required. Global planning allows to give a prior to the local planners and control algorithm, which makes these task tractable. Lap time optimization is specifically developed for this purpose and allows do directly optimize the race line under the constraint of the car. Due to the similarity with predictive control approaches direct shooting approaches to lap time optimization got popularized for autonomous racing in \cite{Vazquez2020}. 
\noindent \textbf{Global Planning}\\
\indent Research from the field of global planning can be roughly divided into different strategies using the objectives of the overall optimization: lap time, geometric properties of the race lines, or energy spent. Racing, as a context for optimization, provides a clear measure of quality in lap time $t_{lap}$ on participating agents. So naturally lowering $t_{lap}$ is a popular choice when it comes to global planning (Figure \ref{fig_global_racelin_comp}).

% do min time papers here
% \cite{Metz1989, Kelly2010, Rucco2015, Gundlach2019, DalBianco2018, Theodosis2012, Pagot2020, Vazquez2020, Herrmann2019, Herrmann2020}
% EA stuff
A first category of global planning approaches is the usage of variations of Evolutionary Algorithms (EAs) to optimize for lap times. This approach is used in \cite{Quadflieg2011, Bevilacqua2017, OKelly2020} with different parameterizations of the search spaces. In these category, an individual ``gene" models a complete configuration of the racing environment, and sometimes vehicle hardware and software. The algorithms also require evaluation functions to gauge the performance of an individual; here, this is the simulated lap time given a configuration. Initially, a pool of genes (referred to as a population) is created by sampling the search space. Then, in each iteration of the algorithm, genes are evaluated, and mutations are performed following different strategies depending on the specific algorithm. Eventually, the individuals in the population should converge, and a best configuration for the global optimal raceline is found. 

Another popular approach when optimizing for lap times as the objective function is to form an optimal control problem (OCP), usually non-linear. The OCP uses the race vehicle's lap times as the objective function, and respects constraints of the geometry and friction limits $\mu_{max}$ of the race track as well as the dynamics and control limits of the racecar. Different proposed approaches usually choose different vehicle dynamic models, and different solvers to solve the optimization problem.
Metz et al.~\cite{Metz1989} formulate an OCP where the objective is the lap time, and solutions are found by quasi-linearization with integral penalty functions, and splicing of constrained and unconstrained arcs to form a two-point boundary value problem.
Kelly et al.~\cite{Kelly2010} use Sequential Quadratic Programming (SQP) to solve the non-linear programming problem where lap time is the objective.
Rucco et al.~\cite{Rucco2015} formulate an OCP where the objective is the lap time. By reformulating the objective with eliminating explicit time step terms, the problem becomes a fixed-horizon free-endpoint problem. Projection operator-based Newton's method is ultimately used for the  trajectory optimization.
Theodosis et al.~\cite{Theodosis2012} initialize the optimization problem with a path created by connecting center line on the straights and clothoids between the center lines. The sequential gradient based, non-linear optimization then uses the lap time as the cost function to find an improved global race line.
Pagot et al.~\cite{Pagot2020} present a non-linear model-predictive framework to formulate a optimal control problem where time is the objective.
Vazquez et al.~\cite{Vazquez2020} formulate a minimum-time optimal control problem by using the centerline of the race track and discretize the continuous space dynamics. Two regularization terms with slip angle cost and a control input rate of change cost are also included in the objective function for reducing the lap time.
Hermann et al.~\cite{Herrmann2019, Herrmann2020, Herrmann2020_2} formulate an optimal control problem where the lap time, the path, the velocity profile and the energy consumption is included in the objective function. By formulating a multi-parametric SQP it is possible to find the optimal velocity plan along a race line while not violating dynamic and energy requirements of the drivetrain.
Finally, the authors of \cite{Lovato2021} devise a apex-finding method and calculates the optimal time global race line by solving an OCP. 

Some researchers show approaches of calculating the race line by satisfying certain geometric properties. Usually, the assumption that vehicles experience lateral acceleration that results in lateral tire forces is made when it comes to autonomous racing. It is then often desirable to minimize the lateral acceleration to minimize the possibility of side slip. In these cases, the aim is to find a race line with the \textit{minimum-curvature} overall. However, it is widely known that the least curvature path is not the ideal path for a racecar which is asymmetric in braking and acceleration and is operated in a combined slip range.
Braghin et al.~\cite{Braghin2008} create design and solve a dynamic problem to find the best compromise between the shortest track and the least curvature track based on the vehicle's dynamic behavior. 
Similarly, Cardamone et al.~\cite{Cardamone2010} also try to find the compromise between the same conflicting objectives, but use a Genetic Algorithm (GA) to find the best weighting parameter between the two objectives.
Heilmeier et al.~\cite{Heilmeier2019} extends the work of Braghin et al.~\cite{Braghin2008} to solve a quadratic optimization problem where constraints on vehicle dynamics limits are set up to find the minimum-curvature path around the race track. This approach is compared to the minimum-time optimization of \cite{Christ2019}. For the same racetrack \cite{Heilmeier2019} achieves a laptime of $t_{lap}$~=~86.13~s while \cite{Christ2019} is even faster with $t_{lap}$~=~84.90~s. The advantage of the minimum-curvature planning is obviously the reduced set of parameters for the vehicle dynamics model and the faster calculation time.

Lastly, some approaches also choose to mimic the geometric properties of a race line driven by human drivers. These approaches often break a turn down into different sections and require race lines to satisfy different properties in different sections. Kuhn~\cite{Kuhn2017} mimics the behavior of a racecar driver by defining the same important decision making points on the track. It calculates the race line by first fixing the locations of the following points: the braking points, turn-in points, apex points, turn-out points, and accelerations points. Then a piecewise rational spline function is used to interpolate all the points and create a race line.
Theodosis et al.~\cite{Theodosis2011} mimic the three phase cornering technique used by professional drivers. It first finds all straights along the track, and linking the straights with curve structures. Then, each connecting curve is found by combining clothoids and a circular arc. The parameters are adjusted to minimize the overall curvature and ensure the path is tangent to the following straight.

\begin{figure}[h]
\begin{center}
\includegraphics{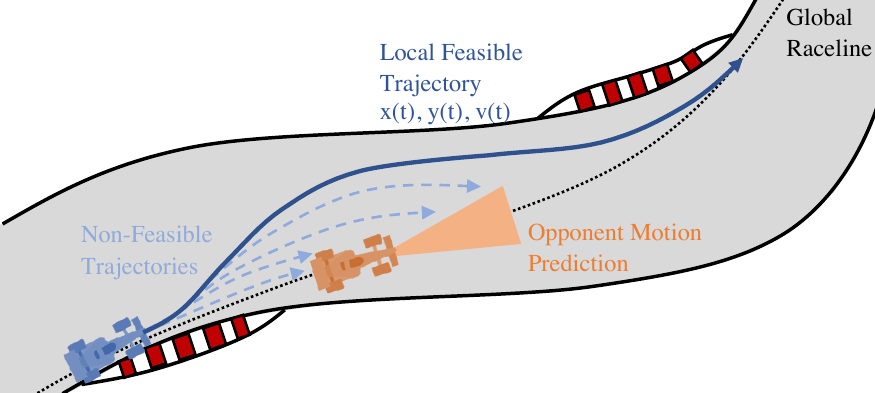}
\caption{Local trajectory planning on the racetrack: The vehicle needs to prediction the opponents motion, plans a feasible trajectory around the opponent and stick closely to the global raceline.}
\label{fig_local_planning}
\end{center}
\end{figure}

\noindent \textbf{Local Planning}\\
\indent In local planning, the main objective is to plan the cars motion for a fixed horizon by avoiding collisions with either the environment or adversaries (Figure \ref{fig_local_planning}). There are three main strategies:
\begin{enumerate}
    \item Modifying the global plan via optimization.
    \item Sample multiple dynamically feasible trajectories and select the best one around obstacles.
    \item Sample in the free space around obstacles to find a feasible trajectory.
\end{enumerate}

From the state of the art we could deduce that many authors use Model Predictive Control (MPC) methods for local trajectory planning. Although this method is control technique, it is also suitable for planning a local trajectory. In this section, we present work that is on the very thin and fluid boundary between planning and control, but which is mainly addressing the planning problem. Pure path tracking with MPC is described later in Subsection \ref{subsec_control}.

% not sure where to put \cite{butz09}
In the first category, the global plan is modified to allow for obstacle avoidance. In these types of formulations, model predictive controllers are usually utilized to optimize the global plan. Upon encountering an obstacle or opponent vehicle, the constraints or cost functions of the optimization problem is modified, and a new motion plan is formulated.
% optimization based stuff
% \cite{ICRAworkshop_08, ICRAworkshop_11, ICRAworkshop_09, Anderson2016, Alcal2020_2, Kapania2016, Funke2017, Subosits2019, Liniger2014, Liniger2015, Williams2016, Vazquez2020, srinivasan2021}
Anderson~et~al.~\cite{Anderson2016} switch between two MPC modes to optimize for minimum-time objective or maximum velocity objective to mimic a professional driver.
Kapania et al.~\cite{Kapania2016} first find the optimal velocity profile given a reference path, and then updates the given path with the fixed velocity profile to find the minimum-curvature path by solving a convex optimization problem.
Williams et al.~\cite{Williams2016} propose a sampling based MPC that relies on path integral control for entropy minimization.
In \cite{Funke2017} a planner is presented that is capable of mediating between conflicting objectives when performing collision avoidance, vehicle stabilization, and path tracking.
Subosits et al.~\cite{Subosits2019} present a real-time trajectory planning algorithm by approximating a re-planning problem as a convex quadratically constrained quadratic problem (QCQP) with a simplified point-mass model.
Alcal et al.~\cite{Alcal2020_2} reformulate the non-linear vehicle dynamics in a Linear Parameter Varying (LPV) form. This can then be used to create a convex optimization problem which is easier to solve for search for a local path.
Kalaria et al.~\cite{ICRAworkshop_08} use a nonlinear-MPC (NMPC) for local planning where the objective consists of progress along the race line, avoiding collision, and use drafting (reduce drag) to make progress.
Brüdigam et al.~\cite{ICRAworkshop_11} use a Gaussian Process (GP) to predict the opponent's maneuver. These stochastic information is then used in a Stochastic MPC to plan efficient overtaking maneuvers.

In the second category, multiple motion primitives, or prototype motion plans, are generated by forward simulating the vehicle dynamics given the current state of the vehicle using multiple different actuation input sequences. This usually results in multiple splines or arcs to select from. In addition, cost functions are used to give each primitive an attached value.  With a search for the best (lowest/highest) cost in these primitives a final trajectory can then be chosen.
% motion primitive stuff here
% \cite{Liniger2014, Liniger2015, Stahl2019_2, OKelly2020, sinha2020}
Liniger et al.~\cite{Liniger2014, Liniger2015} generate a library of trajectories by forward simulating the vehicle using a grid of vehicle velocities and steering angles up to a certain horizon.
Stahl et al.~\cite{Stahl2019_2} propagate the race track with a graph that covers the entire space. The nodes are first placed equidistantly along cross sections of the race track, then edges connecting nodes are created by optimizing for cubic clothoids. This planner was tested on the Roborace vehicle and achieved $v_{max}$~=~223~km/h with an update rate of 16.8~Hz on and NVIDIA Arm electrical control unit (ECU). 
O'Kelly et al.~\cite{OKelly2020} use a uniform grid of points along the global race line as local goal points. Afterwards cubic clothoids are optimized to connect the vehicle's current state and the grid points which leads ultimately to planning a local trajectory.
Finally in \cite{sinha2020} a set of local goal points in front of the vehicle is sampled with the help of a normalizing flow method. Again, cubic clothoids are optimized here to connect the vehicle's current state and the grid points to derive a driveable trajectory.

In the third category, sampling-based methods are used. These approaches randomly sample the free space around the current state of the vehicle for goal states. Once a available goal state is found, a motion plan is then generated connecting the current state of the vehicle with the selected goal state. By introducing randomness in the sampling process, these algorithms are usually efficient, but do not provide guarantees on their optimality.
% rrt, sampling based stuff here
% \cite{Jeong2013, Arslan2017, Feraco2020}
Jeong et al.~\cite{Jeong2013} combine the rapidly-exploring random tree (RRT*) method with a local steering algorithm utilizing the dynamic model of the vehicle.
Arslan et al.~\cite{Arslan2017} combine RRT with closed-loop prediction based on the vehicle model, and incorporate relaxation methods for efficient construction of a tree that guarantees asymptotic optimality.
Feraco et al.~\cite{Feraco2020} combine RRT with Dubins curve to generate dynamically feasible local plans.
Finally, Bulsara et al.~\cite{Bulsara2020} use RRT to find collision free reference paths in the free space. \\
\noindent \textbf{Behavioral Planning}\\
\indent In behavior planning, the focus is usually on high-level decision making on tasks such as selecting an appropriate weighting of different objectives, or selecting plans that impedes the progress of opponents. The research in this area mainly focuses on two different strategies:
\begin{enumerate}
    \item Assigning multiple cost functions with weighting and selecting the plan with the lowest overall cost.
    \item Combine the local planner with game theoretic methods.
\end{enumerate}

In the first category, cost functions are used that represent specific racing values like progress along the track, proximity to the obstacles, effort for control inputs and the deviation from optimal global plan. An overall cost is then found by combining all cost functions for each candidate local trajectory. Cost functions could also incorporate hard constraints by eliminating unqualified plans. Finally, the trajectory with the minimum overall cost is chosen to be the local trajectory.
% cost function stuff here
% \cite{Liniger2014, Liniger2015, OKelly2020, sinha2020}
Liniger et al.~\cite{Liniger2014} use the prototype trajectory without collision and makes the largest progress along the track. This approach is then extend in \cite{Liniger2015} by applying viability kernels on the track which only generates viable trajectories.
O'Kelly et al.~\cite{OKelly2020} assign cost functions representing proximity to the global plan, collision with the environment to prototype trajectories and select the best one.
Finally, Sinha et al.~\cite{sinha2020} assign cost functions for progress along the track, overall curvature, maximum velocity, and collision with predicted opponent motion to all prototype trajectories. \\
\indent In the second category, game theoretic approaches are used to usually find the best action in a two or multiple player game. The continuous motion planning problem is usually transformed into a step-by-step game where each player is allowed to make a ``move" one by one. These approaches usually incorporate the concept of regret to try to find the best response for winning the racing game either immediately or in the long run.
% game theory stuff here
% \cite{Loiacono2010, Williams2017, Notomista2020, wang2020, Liniger2020, Wang2019, Wang2021, Schwarting2021_2, sinha2020}
% Loiacono is kinda weird, not really game theory, could go into the RL section
Williams et al.~\cite{Williams2017} combine a best response model of the opponent behavior with variation of MPC by including predicted opponent trajectories into the cost of other vehicles.
Notomista et al.~\cite{Notomista2020} propose sensitivity-enhanced Nash equilibrium seeking, which uses iterated best response algorithm to optimize for a trajectory in a two car racing game.
In \cite{wang2020} a iterated best response with Nash equilibrium approximation is used to plan receding horizon trajectories. This technique helps to maximally advance the racecar along the track while taking into account opponent's intentions and responses.
Sinha et al.~\cite{sinha2020} build a library of opponent prototypes offline and uses the EXP3 algorithm to solve for a multi-armed bandit problem to approximate the current opponent by using the library.
Liniger et al.~\cite{Liniger2020} repeat the multi-player game in a receding horizon fashion, which results in a sequence of coupled games. With this non-cooperative game approach the authors could show that the vehicles create blocking maneuvers although the risk of a collision gets higher.
Wang et al.~\cite{Wang2021} use sensitivity enhanced iterated best response to seek convergence to the Nash equilibrium in the joint trajectory space for all agents.
Finally, Schwarting et al.~\cite{Schwarting2021_2} use local iterative Dynamic Programming (DP) in belief space to solve a continuous Partially Observable Markov Decision Process (POMDP).

\subsection{Control}
\label{subsec_control}

\begin{figure}[h]
\begin{center}
\includegraphics{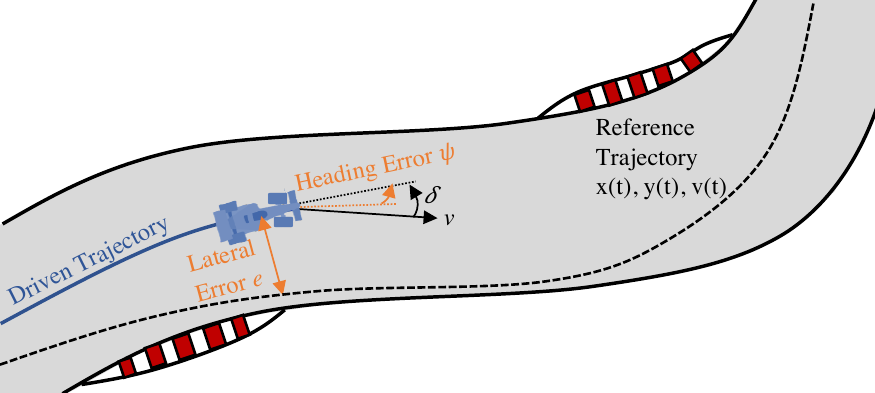}
\caption{Control task in an autonomous racecar. The goal is to reduce lateral and heading error while following a global optimal raceline as a reference trajectory.}
\label{fig_control_overview}
\end{center}
\end{figure}

In the previous subsection, we discussed how to compute either a global or local trajectory on the racetrack. The trajectory includes both a path ($x(t)$, $y(t)$ - position) and velocity profile $v(t)$ which provides the reference information for the \emph{lateral and longitudinal control}. In what follows, we provide control methodologies that leverage such a reference trajectory to compute control actions to navigate the car along the waypoints. The goal is to reduce the lateral and heading error to stay as close to the reference line and to reduce the velocity error to be as fast as possible (Figure \ref{fig_control_overview}). 

At this level of abstraction, the control actions are usually the steering angle $\delta$ and a throttle or brake request (acceleration commands $a_{long}$) that are sent to low-level controllers for actuating the motor and the brakes. We denote $x^\mathrm{ref}(t)$ as the state associated with the reference trajectory at time $t$, and our goal is to design a controller policy $\pi$ that, given the current state of the vehicle at time $t$, denoted as $x(t)$, and the reference state $x^\mathrm{ref}(t)$, computes the control input $u(t)$. 

Table \ref{tab:software_control} gives an overview of the papers that address the control problem for autonomous racing.  The control papers try to address the issue of "handling at the limits" and follow the raceline/reference trajectory as accurate as possible. We define the fundamental problems for autonomous racing control as the following:
\begin{itemize}
    \item High accurate path tracking for low lateral errors.
    \item High accurate path tracking for low heading errors.
    \item High accurate velocity tracking for fast lap times. 
    \item Stable vehicle behavior at high accelerations.
    \item Exact modelling of the nonlinear vehicle behavior.
    \item High control frequency for real-time high speed driving.
\end{itemize}

% \usepackage[normalem]{ulem}
\begin{table*}[h!]
%\begin{tabular}{|r|c|c|c|c|c|}
\label{tab:software_control}
\caption{Overview and categorization of research  paper in the field of control for autonomous racing}
\begin{tabular}{|>{\raggedleft\arraybackslash}m{3.20cm}|>{\centering\arraybackslash}m{1.3cm}|>{\centering\arraybackslash}m{3.0cm}|>{\centering\arraybackslash}m{5.5cm}|>{\centering\arraybackslash}m{1.3cm}|>{\centering\arraybackslash}m{1.15cm}|}

\hline
\textbf{Name and Reference} & \textbf{Year} & \textbf{Control Category}          & \textbf{Method}                     & \textbf{Tested on Hardware} & \textbf{Racing Series} \\ \hline

Voser et al.  \cite{Voser2010}                          & 2010          & Classic Control           & Controller Analysis                           & Yes       & -              \\ \hline
Kritayakirana et al.~\cite{Kritayakirana2010}           & 2010          & Classic Control           & Feedforward Longitudinal Controller           & Yes       & -              \\ \hline
Talvala et al.  \cite{Talvala2011}                      & 2011          & Classic Control           & Lookahead control                             & Yes       & -              \\ \hline
Kritayakirana et al.~\cite{Kritayakirana2012, Kritayakirana2012_2} & 2012  & Classic Control        & COP feedforward and feedback steering         & Yes       & -              \\ \hline
Kapania   et al.~\cite{Kapania2015}                     & 2015          & Classic Control           & Feedback-feedforward steering controller      & Yes       & -              \\ \hline
Park  et al.~\cite{Park2015}                            & 2015          & Classic Control           & Convex optimization                           & Yes       & -              \\ \hline
Laurense et al.~\cite{Laurense2017, Laurense2018}       & 2017, 2018     & Classic Control          & Slip angle-based control strategy             & Yes       & -              \\ \hline
Ni et al.~\cite{Ni2017, Ni2019}                         & 2017, 2019     & Classic Control          & Controller Framework                          & Yes       & FSD            \\ \hline
Fu et al.~\cite{Fu2018}                                 & 2018          & Classic Control           & Maximize GG Diagram                           & Yes       & FSD            \\ \hline
Chatzikomis et al.~\cite{Chatzikomis2018}               & 2018          & Classic Control           & Torque Vectoring                              & Yes       & Roborace       \\ \hline
Wachter et al.~\cite{Wachter2020}                       & 2020          & Classic Control           & Controller Analysis                           & No        & -               \\ \hline
Pedone et al.~\cite{Pedone2020}                         & 2020          & Classic Control           & Model-free nonlinear control                  & No        & -               \\ \hline
Sukhil et al.~\cite{Sukhil2021}                         & 2021          & Classic Control           & Adaptive Lookahead for Pure Pursuit           & Yes       & F1TENTH         \\ \hline
Beal et al.~\cite{Beal2012}                             & 2012          & Model Predictive Control  & Model Predictive Envelope Control             & Yes       & -               \\ \hline
Williams et al.~\cite{Williams2016}                     & 2015          & Model Predictive Control  & Model Predictive Path Integral Control (MPPI) & No        & -               \\ \hline
Carrau et al.~\cite{Carrau2016}                         & 2016          & Model Predictive Control  & sparse Randomized MPC         & Yes       & 1:43 car                       \\ \hline
Verschueren et al.~\cite{Verschueren2016}               & 2016          & Model Predictive Control  & Nonlinear MPC (NMPC)                          & No        & -              \\ \hline
Drews et al.~\cite{drews2017}                           & 2017          & Model Predictive Control  & MPPI + CNN                                    & No        & -              \\ \hline
Liniger et al.~\cite{Liniger2017}                       & 2017          & Model Predictive Control  & sparse Randomized MPC                         & No        & 1:43 car       \\ \hline
Williams et al.~\cite{Williams2017, Williams2018, Williams2018_2, Williams2018_3}  & 2017, 2018   & Model Predictive Control   & MPC+ Game Theory, MPC + RL, Sampling based MPC  & No & AutoRally \\ \hline
Novi et al.~\cite{Novi2019}                             & 2019          & Model Predictive Control  & Hierarchial Nonlinear MPC (NMPC)              & No        & -                 \\ \hline
Liniger et al.~\cite{Liniger2019}                       & 2019          & Model Predictive Control  & MPC + Viability Theory                        & Yes       & 1:43              \\ \hline
Brown et al.~\cite{Brown2020}                           & 2020          & Model Predictive Control  & Nonlinear MPC (NMPC)                          & Yes       & -                 \\ \hline
Liu et al.~\cite{Liu2020}                               & 2020          & Model Predictive Control  & Standard MPC                                  & No        & FSD               \\ \hline
Alcal et al.~\cite{Alcal2020}                           & 2020          & Model Predictive Control  & Linear Parameter Varying MPC (LPV-MPC)        & Yes       & -                 \\ \hline
Gandhi et al.~\cite{Gandhi2021}                         & 2021          & Model Predictive Control  & Robust MPPI (RMPPI)                           & No        & AutoRally        \\ \hline
Pour et al.~\cite{Pour2021}                             & 2021          & Model Predictive Control  & Linear Parameter Varying MPC (LPV-MPC)        & No        & -                 \\ \hline
Wischnewski et al.~\cite{Wischnewski2021}               & 2021          & Model Predictive Control  & Tube MPC (TMPC)                               & No        & Roborace          \\ \hline
Li et al.~\cite{ICRAworkshop_02}                        & 2021          & Model Predictive Control  & NMPC + MIQP                                   & No        & -                 \\ \hline
O Kelly et al.~\cite{OKelly2020}                        & 2020          & Optimization              & CMA-ES                                        & No        & F1TENTH           \\ \hline
Kapania et al.~\cite{Kapania2015_2}                     & 2015          & Learning Based Control    & PD-ILC, Q-ILC                                 & Yes       & -                 \\ \hline
Brunner et al.~\cite{Brunner2017}                       & 2017          & Learning Based Control    & Learning MPC (LMPC)                           & No        & 1:10              \\ \hline
Rosolia et al.\cite{Rosolia2017,Rosolia2017_2,Rosolia2017_3,Rosolia2020}    & 2017, 2020  & Learning Based Control &  Learning MPC (LMPC)           & No        & -                 \\ \hline
Ji et al.~\cite{Ji2018}                                 & 2018          & Learning Based Control     &  DNN + backstepping variable structure control & Yes     & FSD               \\ \hline
Bujarbaruah et al.~\cite{Bujarbaruah2018}               & 2018          & Learning Based Control     &  Adaptive MPC                                & No        & -                 \\ \hline
Hewig et al.~\cite{Hewing2018}                          & 2018          & Learning Based Control     &  NMPC + Gaussian Process                     & No        & 1:43 car          \\ \hline
Wagener et al.~\cite{wagener2019}                       & 2019          & Learning Based Control     &  Dynamic Mirror Descent MPC (DMD-MPC)        & No        & AutoRally        \\ \hline
Kabzan et al.~\cite{Kabzan2019}                         & 2019          & Learning Based Control     &  NMPC + Gaussian Process                     & Yes       & FSD               \\ \hline
Wischnewski et al.~\cite{Wischnewski2019_2} \cite{Wischnewski2020} & 2019 & Learning Based Control   & Controller + Gaussian Process; recursive Least-Mean-Squares algorithm & Yes & Roborace \\ \hline
Vallon et al.~\cite{Vallon2020}                         & 2020          & Learning Based Control     & Learning MPC (LMPC)                          & No        & -                 \\ \hline
Jain et al.~\cite{jain2020}                             & 2020          & Learning Based Control     & MPC + Gaussian Process                       & No        & F1TENTH           \\ \hline
Kapania et al.~\cite{Kapania2020}                       & 2020          & Learning Based Control     & PD-ILC, Q-ILC                                & Yes       & -                 \\ \hline
Van Niekerk et al.~\cite{vanniekerk2020online}          & 2020          & Learning Based Control     & Receding Horizon Control + Gaussian Processes& No        & -                 \\ \hline
Xiao et al.~\cite{ICRAworkshop_07}                      & 2021          & Learning Based Control     & Controller DNN +                             & No        & F1TENTH           \\ \hline
Hindiyeh et al.~\cite{Hindiyeh2014}                     & 2014          & Drifting Control           & Controller Framework                         & Yes       & -                 \\ \hline
Goh Jet al.~\cite{Goh2016, Goh2018, Goh2019}            & 2016-2019     & Drifting Control           & Controller Framework                         & Yes       & -                 \\ \hline
Zubov et al.~\cite{Zubov2018}                           & 2019          & Drifting Control           & Controller Framework                         & No        & Roborace          \\ \hline
Joa et al.~\cite{Joa2020}                               & 2020          & Drifting Control           & Controller Framework                         & No        & -                 \\ \hline
Cai et al.~\cite{Cai2020}                               & 2020          & Drifting Control           & Deep Learning                                & No        & -                 \\ \hline
\end{tabular}
\end{table*}

\noindent For a better overview and understanding of the different types of control research we decided to split the papers into six subsections.

\noindent \textbf{Classic Control}\\
\indent Firstly, we survey papers that cover classical and well-known principles in the field of path and velocity tracking. Ni and Hu \cite{Ni2017} present a path following controller for their FSD vehicle. Their overall controller architecture consists of longitudinal, lateral and yaw controllers that operate the vehicle on a predefined G-G diagramm (maximum lateral and longitudinal acceleration.) In both \cite{Talvala2011} and \cite{Sukhil2021} we see the usage of a simple lookahead controller that provides path tracking at high speeds event at the limits of tire adhesion. In \cite{Kritayakirana2012} an autonomous racing controller is presented that uses the vehicle's centre of percussion (COP) to design a feedforward and feedback steering. This showed how to simplify the equations of motion and highlights the challenge of controlling a vehicle with highly saturated tires. Furthermore, a special focus on path tracking at the tire/vehicle limits is presented in the work of \cite{Kritayakirana2012_2,Kapania2015, Laurense2017,Fu2018}. While the research of \cite{Kritayakirana2012_2} and \cite{Fu2018} displays the usage of a G-G diagram to display controllers that operate the vehicle at the limits, Laurense et al.~\cite{Laurense2018} presents a slip angle-based control strategy to maintain the front tires at a certain slip angle to create the maximum tire forces. A special focus on longitudinal control (speed control) is shown in \cite{Kritayakirana2010,Laurense2017,Pedone2020}. Laurense et al.~\cite{Laurense2017} presents a control framework for  full tire-force utilization with slip-angle based steering control, combined with explicit control of the path-tracking dynamics through longitudinal speed feedback to achieve a better path tracking. In \cite{Pedone2020} a model-free nonlinear controller for longitudinal speed control is presented. In this approach, a dynamic reconstruction of information on the vehicle’s motion concerning the inputs acting on the system with sensor data is displayed. With this its possible to reconstruct the maximum longitudinal tire forces for current states which can be used for accurate speed tracking. Finally, in both \cite{Voser2010} and \cite{Wachter2020} additional sensitivity analysis of path controlling at the limits and high sideslip maneuvers are displayed.\\
\noindent \textbf{Model Predictive Control}\\
\indent The second and most-popular strategy used to synthesis autonomous racing controller is \textit{Model Predictive Control}. In MPC a sequence of control actions is computed by forecasting the future trajectory of the vehicle over a short time window. In particular, given the state of the system $x_t$, an MPC solves a \textit{Finite Time Optimal Control Problem (FTOCP)} to compute an optimal sequence of states $\{x_t^*, \ldots, x_{t+N}^*\}$ and inputs $\{u_t^*, \ldots, u_{t+N-1}^*\}$  over a fixed horizon $N$. In autonomous racing the objective of the optimal control problem is to either track a global reference trajectory or to minimize the lap time. Upon computing such sequence of optimal states and actions, the first control action $u_t^*$ is applied to the system and the process is repeated at the next time step based on the updated state $x_{t+1}$. MPC-based methodologies are the main method behind several autonomous racing controller which have been implemented on real vehicles. The advantages of MPC are that $1)$ forecast is used to act proactively and to $2)$ feedback is naturally incorporated in the controller that repeatedly updates the optimal trajectory. Notice that when the planning horizon $N$ is short, the planned trajectory may not account for the future behavior of the system and as a result the controller may take shortsighted control actions. However, computing such quantities that exactly approximate the cost and constraint beyond the prediction is challenging. In practical applications it is preferred either to use a long prediction horizon~\cite{Liniger2017} or to approximate these quantities based on historical data~\cite{Rosolia2020}.

% Throughout this section, we denote $x_t$ and $u_t$ as the predicted state and input, respectively. Our goal is to compute a sequence of $N-1$ optimal actions $\{u_t^*, \ldots, u_{t+N-1}^*\}$ such that the deviation between the predicted trajectory $\{x_t^*, \ldots, x_{t+N}^*\}$ and the reference trajectory $\{x^{\mathrm{ref}}(t), \ldots, x^\mathrm{ref}(t+N)\}$ is minimized. Thus, this is designed as a \textit{Finite Time Optimal Control Problem (FTOCP)}.
% In the above FTOCP, the planned trajectory is initialized at the current state of the vehicle $x(t)$, and the sets $\mathcal{X}$ and  $\mathcal{U}$ represent the state constraints (e.g. the lane boundaries) and the input saturation constraints. Furthermore, we assumed that the reference trajectory is feasible for the used dynamic vehicle model.
% The control action from the MPC is a function of the current state of the system $x(t)$, which is used to initialize the predicted trajectory. To avoid this issue, in the FTOCP the terminal cost $V(x_{t+N})$ and terminal constraint $\mathcal{X}_F$ are used to approximate the tail of the cost and constraint beyond the prediction horizon. 

\begin{figure}[h]
\begin{center}
\includegraphics[scale=0.97]{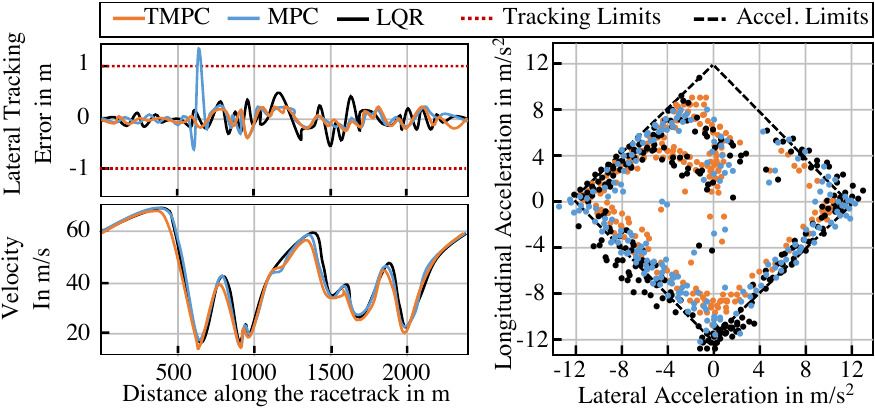}
\caption{Qualitative comparison of an LQR, MPC and Tube MPC controller in an autonomous racing setup based on \cite{Wischnewski2021}. While the MPC is outperforming the LQR controller, it can be seen the the Tube-MPC is not violating any dynamical constraints e.g. the maximum lateral and longitudinal acceleration.}
\label{fig_control_comp}
\end{center}
\end{figure}

In \cite{Williams2016} a sampling based MPC algorithm is derived. This so called \textit{Model Predictive Path Integral Control (MPPI)} algorithm is using the methodology of path integral control that derives an optimal control based on stochastic sampling of trajectories.  It is demonstrated that this approach explicitly provides a formula for the controls over the entire time horizon and that it relaxes the usual condition between control authority and noise required in path integral control. The authors use this fundamental control approach and enhance it with additional decision maker \cite{Gandhi2021},  game theory \cite{Williams2018}, DNNs \cite{drews2017} and reinforcement learning \cite{Williams2017_2} methods to derive further improvements. 

Carrau et al.~\cite{Carrau2016} present at \textit{sparse Randomized MPC} (SRMPC) approach that is based on a \textit{Stochastic MPC}. This approach is used to deal with model uncertainty at high speeds and high accelerations. While driving with the vehicle it collects data along the track which is then used to identify the model uncertainty probabilistically. This tightens the constraints for the MPC automatically while still having the size and structure of a standard MPC problem. The optimization problem is solved in 20~ms for a 1:43 and the results show that for a desired violation the controller achieves faster lap times and fewer constraints violations than a standard MPC algorithm. \cite{Liniger2017} is building on top of this approach and enhancing it with disturbance feedback policies to optimize over the state feedback matrices. \\
\indent In order to capture the nonlinear dynamics of the vehicle and tires a \textit{nonlinear MPC} (NMPC) can be designed and modelled \cite{Brown2020}. Although this is computationally expensive, with the help of nonlinear optimization solver like FORCES PRO \cite{FORCESPro} these type of optimizations can be solved in real-time on the vehicle. In \cite{Novi2019} a  \textit{hierarchical NMPC} is presented that consists of two controllers: Firstly, a high-level NMPC with point-mass model  that simplifies the vehicle dynamics and is constraint by the tire G–G diagram. Secondly, a low-level NMPC with a high-fidelity model uses the output (velocity profile) of the first NMPC as a terminal constraint. This method helps to reduce the prediction horizon and therefore calculate the vehicle dynamics in real-time. Furthermore, this approach was improved in \cite{Vazquez2020} with a simpler vehicle model to run on a FSD vehicle.

Li et al. \cite{ICRAworkshop_02} use a NMPC with a minimum-time objective and a collision-avoidance constraint. By applying a Mixed Integer Quadratic Programming (MIQP) method a control strategy is created that is optimized regarding the safety and the laptime. The authors conclude that for such an approach the prediction horizon needs to be large enough for creating feasible results although this leads to a higher computation time and is therefore non real-time capable. The authors of \cite{Liniger2019} combine a low-level MPC with a viability kernel that  efficiently generates finite look-ahead trajectories to maximize the progress of the car.  At the same time the viability kernel creates trajectories that  remaining recursively feasible with respect to static obstacles. The authors apply this algorithm to both a simulation where the effects of various design choices and parameters are identified and showed that a hierarchical controller can be improved by incorporating the viability kernel in the trajectory planning phase. 

Beal et al.~\cite{Beal2012} use estimations of the friction coefficient and vehicle sideslip to define state constraint and unstable vehicle behaviors. This information is utilized in an model predictive envelope controller to create a region of stable vehicle motions. With this approach it is possible to operate the vehicle on the handling and stability limits. Similarly, Wischnewski et al \cite{Wischnewski2021} and Williams et al.~\cite{Williams2018_3} present a \textit{Tube-MPC} (TMPC) approach where nonlinear effects and external disturbances are taken into account of the MPC design. By approximating a tube of reachable sets over the prediction horizon the vehicle guarantees a space of constraint satisfaction.
Finally, because tuning the parameters of a controller is time-consuming and needs experience from experts some researches try to automatically tune the parameters with optimization techniques. In \cite{OKelly2020} the superoptimization toolchain \emph{TUNERCAR} is presented which is using a Covariance Matrix Adaptation Evolution Strategy (CMA-ES) to optimize both vehicle hardware parameters (center of gravity, mass) and control parameters (P-, I-, D-parameters) of the car. The evaluation is based on the laptime of the car and the algorithm shows the capability of reducing the laptime driven by the car by optimizing these parameters.

\begin{figure}[h]
\begin{center}
\includegraphics[scale=1.08]{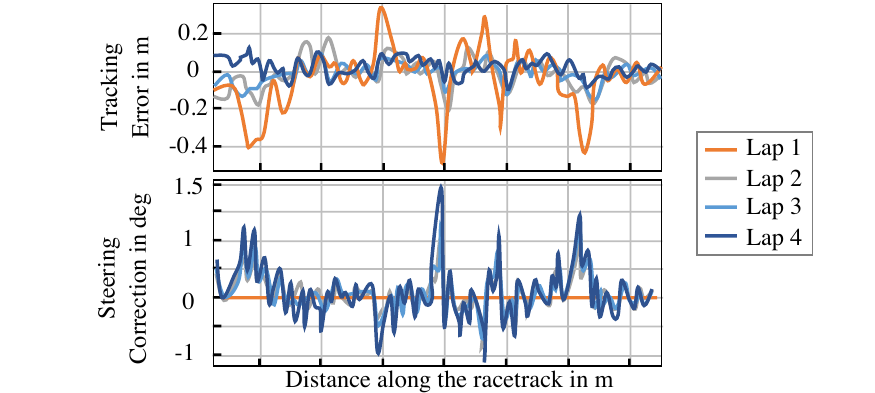}
\caption{Qualitative demonstration of a learning-based control approach based on \cite{Kapania2015_2}. For each new lap the car is decreasing the tracking error by applying a higher correction steering angle for each vehicle movement.}
\label{fig_learning_control}
\end{center}
\end{figure}

\noindent \textbf{Learning Based Control}\\
\indent An additional control strategy to improve the tracking of a reference trajectory is to leverage \textit{Iterative Learning Control (ILC)} based methodologies~\cite{bristow2006survey}. ILC methods are useful for applying them to an autonomous race vehicle, since the vehicle is running on the race track repeatedly for multiple laps. In this case the tracking error and vehicle data from previous laps can be used to compute a feedforward correction term that improves the path and velocity tracking performance significantly (Figure \ref{fig_learning_control}). ILC-based strategies for autonomous racing have been successfully implemented on full-size vehicles~\cite{Kapania2015_2,Kapania2020}. 

In \cite{Brunner2017,Rosolia2017,Rosolia2017_2,Rosolia2017_3, Rosolia2020} a  \textit{Learning MPC (LMPC)} is proposed which is an optimization and data-driven framework to make the car faster every lap and therefore reduce the laptime. The optimal control problem of the MPC is enhanced in a way so it tries to compute a solution by solving at time $t$ of each lap the finite time constrained optimal control problem. This creates a convex optimization problem which can be solved with respective solvers. The authors show that the LMPC finds a faster trajectory for each new lap while maintaining the set constraints.

Hewing et al.~\cite{Hewing2018} present an \textit{learning-based cautious NMPC} which aims to learn from vehicle sensor data with Gaussian Processes to improve the vehicle dynamics model. The GP model is used for regression to  identify uncertainties and a mismatch in the vehicle dynamics model parameters based on measurement data. The NMPC is extended with this learning model and reformulated in a stochastic setting which improves the performance and safety of the vehicle at the same time. Furthermore, the authors implement this approach on a FSD vehicle in \cite{Kabzan2019} and demonstrate the implementation and experimental validation of this kind of learning-based control approach. Finally, Jain et al.~\cite{jain2020} were using the same approach but only with an extended kinematic vehicle model for the MPC to proof, that this type of learning-based control can also leverage the usage of simplified vehicle models.

Because of the control system quality and unmodelled effects, it is well known that there is a gap between the planned and the driven trajectory. This gap is unknown and depended on the environment the vehicle is driving in. To mitigate this gap, \cite{Wischnewski2019_2} presents a learning control approach on the method of Gaussian Process for a nonlinear regression. This GP learns online, while driving, how big this gap is and then tries close it over the time by using a so called scale-factor. This scale-factor serves as an optimization variable that tries to maximize longitudinal and lateral accelerations each lap.

Finally, the authors of \cite{Ji2018} proposed a control scheme that consists of a robust steering controller and a DNN. While the path tracking is done via \textit{backstepping variable structure control (BVSC)} the DNN is integrated to estimate nonlinear functions, e.g. the uncertainty of tire cornering stiffness. 

\noindent \textbf{Drifting Control}\\
\indent Although it is not following an optimal raceline, racing head-to-head or striving for the fastest laptime, the field of autonomous drifting is a special subcategory for autonomous racing. Here, the researchers show algorithms that are able to maneuver the car autonomously beyond the stable handling of limits and stabilize the car in a point of high slip angle. In \cite{Hindiyeh2014} a successive loop structure is presented as a controller that tracks only the vehicles sideslip based on the yaw rate as a control input. Goh et al.~\cite{Goh2016,Goh2018,Goh2019} present a controller framework that is able to drift autonomously with the vehicle while tracking a predefined reference path. This enables drifting maneuvers at special references path, e.g. a circle or figure of 8. The authors using a single track vehicle model \cite{Goh2016} and experiment with different variations of control values (sideslip \cite{Goh2016}, rotation of the vehicle’s velocity vector to track the lateral error \cite{Goh2018,Goh2019}) to reach a sideslip angle of up to -40 degree with speeds up to 45 km/h on a real test vehicle. Finally, Joa et al.~\cite{Joa2020} present a 3-level structure for a drift controller. Firstly, they designed a supervisor which determines rate and rear longitudinal slip ratio for the drift maneuver. Secondly, a upper-level controller calculates lateral force (front) and longitudinal force (rear) for tracking the planned vehicle motion. Finally, a low-level controller converts the commands (forces) defined by the  upper-level controller into control inputs for the vehicles (throttle, steering angle). With this setup the authors derived a steady state drifting on a real test vehicle.

\subsection{End-to-End}
\label{subsec_end2end}
\begin{figure}[h]
\begin{center}
\includegraphics{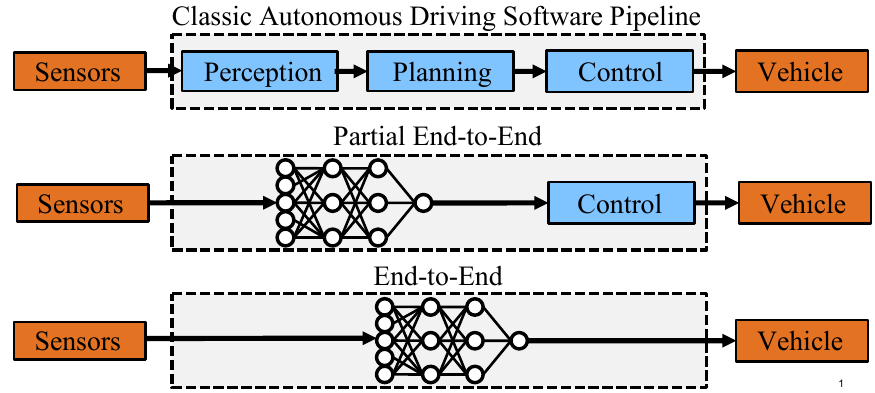}
\caption{Classic autonomous driving software pipeline in comparison to partial and full end-to-end software pipeline.}
\label{fig_end2end}
\end{center}
\end{figure}

The previous subsections described in detail which research efforts in the fields of perception, planning and control have been achieved for autonomous racing. Besides that, researchers have tried to set up partial or full \textit{end-to-end} approaches to master the autonomous racing task. As displayed in  Figure \ref{fig_end2end}, in the context of autonomous driving end-to-end means that either partial modules or all software modules are completely replaced with data-driven approaches like a DNN. 

The partial end-to-end approach aims towards replacing or combining modules with a DNN. This has the advantage that the DNN provides a low-dimensional intermediate representation of the racetrack (e.g. a trajectory) that can be then used in a classic control systems (e.g. PID-Controller). In contrast to a full end-to-end system the final actuator output (steering angle, throttle position) is not predicted directly. On the one hand the field of autonomous racing provides a perfect proving ground for end-to-end approaches: clear driveable area, no signage (e.g. traffic lights), one class of objects, clear objective for training (fastest laptime). On the other hand, when using end-to-end systems large data sets are needed to train the DNNs. This data must contain a wide variety of situations so that the DNN achieves a good level of performance. Similar to other DNN applications the generalization and performance of these systems are the biggest issues. In addition to these common known issues, we define the following main problems for end-to-end autonomous racing:
\begin{itemize}
    \item Open to question system architecture design: partial vs. full end-to-end.
    \item Difficulty to learn the vehicle dynamics parameter - especially the nonlinear vehicle and tire dynamics.
    \item Training purely in simulation environments lead to \textit{simulation-to-reality} gap. 
    \item High amount of various data necessary for training the artificial networks.
    \item Out of distribution events can cause drastic failure cases: Driving at high speeds is rare and thus learning how to correctly react is difficult
\end{itemize}

In the following we present research efforts that use end-to-end approaches for autonomous racing which are summarized in categories in Table \ref{tab:software_EndtoEnd}.

\begin{table*}[]
\centering
\caption{Overview and categorization of additional software in the field of end-to-end approaches for autonomous racing}
\label{tab:software_EndtoEnd}
\begin{tabular}{|>{\raggedleft\arraybackslash}m{2.8cm}|>{\centering\arraybackslash}m{1.3cm}|>{\centering\arraybackslash}m{2.8cm}|>{\centering\arraybackslash}m{2.7cm}|>{\centering\arraybackslash}m{2.8cm}|>{\centering\arraybackslash}m{1.2cm}|>{\centering\arraybackslash}m{1.5cm}|}
\hline
\textbf{Name and Reference} & \textbf{Year} & \textbf{End-to-End Category} & \textbf{Topic} & \textbf{Method} & \textbf{\begin{tabular}[c]{@{}c@{}}Tested on\\ Hardware\end{tabular}} & \textbf{Racing Series}  \\ \hline

Perez et al.~\cite{Perez2008}                       & 2008          & Optimization              & Optimal Control Policy                & Evolutionary Algorithm    & No        & - \\ \hline
Salem et al.~\cite{Salem2017, Salem2018, Salem2019} & 2017, 2018    & Optimization              & Optimal Control Policy                & Fuzzy Logic               & No        & -  \\ \hline
Korkmaz et al.~\cite{Korkmaz2018}                   & 2018          & Optimization              & Optimal Control Policy                & Fuzzy Logic               & No        & -  \\ \hline
Oliveira  et al.~\cite{Oliveira2018}                & 2018          & Optimization              & Vision based Planning                 & Bayesian Optimization     & No        & - \\ \hline
Lee et al.~\cite{Lee2019}                           & 2019          & Deep Learning             & Vision based Planning                 &  MPC + CNN                & Yes       & AutoRally \\ \hline
Weiss et al.~\cite{weiss2020,Weiss2020_1}           & 2020          & Deep Learning             & Vision based Planning                 & CNN, RNN                  & No        & - \\ \hline
Tatulea et al.~\cite{Tatulea2020}                   & 2020          & Deep Learning             & trajectory planning                         & NMPC \& DNN               & No        & F1TENTH \\ \hline
Weiss et al.~\cite{ICRAworkshop_05}                 & 2021          & Deep Learning             & Trajectory Prediction                 & RNN                       & No        & - \\ \hline
Drews et al.~\cite{Drews2019}                       & 2019          & Deep Learning             & Localization                          & CNN, LSTM, + MPC          & Yes       & AutoRally \\ \hline
Mahmoud et al.~\cite{Mahmoud2020}                   & 2020          & Deep Learning             & Vision based Planning                 & CNN, LSTM                 & Yes       & Donkey Car \\ \hline
Wadeka et al.~\cite{ICRAworkshop_10}                & 2021          & Deep Learning             & Vision based Planning                 & CNN                       & No        & IAC \\ \hline
Perot et al.~\cite{Perot2017}                       & 2017          & Reinforcement Learning    & Vision based Planning                 & Advantage actor-critic    & No        & - \\ \hline
Jaritz et al.~\cite{Jaritz2018}                     & 2018          & Reinforcement Learning    & Vision based Planning                 & Advantage actor-critic    & No        & - \\ \hline
De Bruin et al.~\cite{deBruin2018}                  & 2018          & Reinforcement Learning    & Vision based Planning                 & Q-Learning+ State representation Learning & No & - \\ \hline
Remonda et al.~\cite{Remonda2019}                   & 2019          & Reinforcement Learning    & Vision based Planning                 & DDPG                      & No        & - \\ \hline
Niu et al.~\cite{Niu2020}                           & 2020          & Reinforcement Learning    & Vision based Planning                 & DDPG                      & No        & - \\ \hline
Gückiran et al.~\cite{Guckiran2019}                 & 2019          & Reinforcement Learning    & Vision based Planning                 & SAC, Rainbow DQN          & No        & - \\ \hline
Fuchs et al.~\cite{Fuchs2021}                       & 2021          & Reinforcement Learning    & Vision based Planning                 & SAC                       & No        & - \\ \hline
Chisari et al.~\cite{chisari2021}                   & 2021          & Reinforcement Learning    & Vision based Planning                 & SAC + policy output regularization    & Yes & 1:43 car \\ \hline
Lee et al.~\cite{Lee2020}                           & 2021          & Reinforcement Learning    & Vision based Planning                 & Bayesian Deciscion Making & Yes       & AutoRally \\ \hline
Pan et al.~\cite{Pan2018}                           & 2021          & Reinforcement Learning    & Vision based Planning                 & Imitation Learning        & Yes       & AutoRally \\ \hline
Cai et al.~\cite{Cai2021}                           & 2021          & Reinforcement Learning    & Vision based Planning                 & Imitation Learning        & Yes       & 1:20 car \\ \hline
Schwarting et al.~\cite{schwarting2021}             & 2021          & Reinforcement Learning    & Vision based Planning                 & Model based RL            & No        & - \\ \hline
Brunnbauer et al.~\cite{Brunnbauer2021}             & 2021          & Reinforcement Learning    & Vision based Planning                 & Model based RL            & Yes       & F1TENTH\\ \hline
Song et al.~\cite{song2021}                         & 2021          & Reinforcement Learning    & Vision based Planning + Overtaking    & SAC + 3-stage curriculum learning     & No & - \\ \hline
Gundu et al.~\cite{Gundu2019}                       & 2019          & Reinforcement Learning    & Model-free optimal control            & Q Learning + Soft-Actor Critic        & No     & - \\ \hline
Ivanov et al.~\cite{Ivanov2020}                     & 2020          & Reinforcement Learning    & Verification                          & Variation of Algorithms   & Yes       & F1TENTH \\ \hline

\end{tabular}
\end{table*}

Perez et al.~\cite{Perez2008} derives the control commands based on a rule-based evolutionary strategy. Although the car is driving successfully around the racetrack and follows a raceline the controller is only able to handle low speeds. In both \cite{Salem2017,Salem2019} the authors propose two fuzzy controllers for calculating the steering angle and computing the target speed of the car based on sensor information in the TORCS simulator \cite{Torcs2005}. In Olivera et al.~\cite{Oliveira2018} Bayesian optimization (BO) is used to find a control policy that minimises the time per lap while keeping the vehicle on the racetrack. The BO helps to search more efficiently over high-dimensional policy-parameter spaces an outperforms other evolutionary algorithms.\\
 \indent A solution for a partial end-to-end approach is presented in  \cite{weiss2020,Weiss2020_1} and \cite{ICRAworkshop_05}. The \textit{DeepRacing Framework} is an end-to-end simulation environment and virtual testbed for training and evaluating algorithms especially for autonomous racing. In \cite{Weiss2020_1} three versions to control the racecar are presented: Pixel to control, pixel to waypoints, pixel to curves. It was shown that a partial end-to-end approach that provides parameterized trajectories based on a DNN outperforms a full end-to-end approach in terms of laptime and failures. \cite{Lee2019} shows the combination of MPC and CNNs to create a \textit{ perceptual attention-based predictive control algorithm}. With this, the MPC learns how to place attention on relevant areas of a visual input, which allows the vehicle to detect unsafe conditions faster. Drews at al. \cite{Drews2019} are providing a framework that combines DNN based road detection as well as MPC to drive aggressively using only the sensor data from a monocular camera, IMU, and wheel speed sensors on the AutoRally vehicle. By combining CNNs and a Long Short Term Memory (LSTM) network the car is able to learn a local cost map representation of the track based on the camera input. This enables a global position estimation with a particle filter against a schematic map at high speeds. The local trajectory planning is afterwards done with the MPC. The authors of \cite{Mahmoud2020} provide an evaluation about the image sizes for a full end-to-end approach on a 1:10 scale vehicle. Based on their experiments the authors show that by decreasing the image size as an input for the end-to-end pipeline the car can drive faster and has a higher response time. Another evaluation for the usage of end-to-end algorithms is done by Wadekar~et~al.~\cite{ICRAworkshop_10} in an simulation environment. The authors explored different data collection strategies with the goal of reaching high speeds and stable driving with the racecar. They conclude that even in the racetrack setup a high diversity and high amount of training data is necessary to achieve decent results. 
 
\indent Beside these plain usage of DNNs in the software pipeline, additional research efforts were done in the field of reinforcement learning. The autonomous racecar is seen as an agent that interacts with its environment in a continuous form. At each timestep $t$ the agent fulfills an action $a_t$ that leads to a reward $r_t$ as well as an observations of all environment states $s_t$. Based on the reward $r_t$ the agent tries to maximize the sum of the rewards over time and therefore can \textit{learn} a specific behavior in this environment.  Autonomous racing researchers that develop RL algorithms are using either the F1TENTH Gym \cite{okelly2019}, the Roborace Simulator \cite{herman2021}, the SVL Simulator \cite{Guodong2020} or TORCS \cite{Torcs2005}. All these simulation environments have an openAI Gym \cite{openAI2016} interface that was created for the setup of RL developments. Both \cite{Perot2017} and \cite{Jaritz2018} address the problem of autonomous racing by applying the method of Advantage Actor-Critic (A3C) to a simulation rally racing game. A complete framework for the training of the autonomous agents in a distributed system with different tracks and road conditions is presented as well as the RL method for achieving the end-to-end driving. They generate reasonable results with a fast and reliable vehicle maneuver, especially on different road conditions, but the approach fails to generalize well. To create a better generalization for different racetracks de Bruin et al.~\cite{deBruin2018} provide a combination of Q-Learning and state representation learning to display that this combination learns policies quicker and generalize better to new racetracks then single RL. Both \cite{Remonda2019} and \cite{Niu2020} use the method of Deep Deterministic Policy Gradient (DDPG) to explore the usage of RL in autonomous racing in the TORCS simulation. In both experimental setups DDPG is specially enhanced for the usage on the racetrack and shows good learning and execution results. The application of Soft Actor Critic (SAC) with enhancement and variations is displayed in \cite{Guckiran2019,song2021,chisari2021}. While \cite{Guckiran2019} is only using a simulation, Chisari et al.~\cite{chisari2021} are applying this approach to 1:43 small-scale vehicles and compare the SAC method to a MPC path planner -- the MPC outperforms the RL method. The work from Fuchs and Song et al.~\cite{Fuchs2021,song2021} is using the racing game \textit{Gran Turismo} as both a training and evaluation environment. While in \cite{Fuchs2021} the framework for training the RL agents is presented Song et al.~\cite{song2021} uses and enhances this approach not only to drive with a single vehicle but also with multiple agents. They show that their RL agent is capable of driving fast, following the raceline and overtaking other agents without crashing. Finally, Schwarting et al.~\cite{schwarting2021} and  Brunnbauer et al.~\cite{Brunnbauer2021} present model-based reinforcement learning approaches which can learn competitive visual control policies through self-play in imagination (World Models idea \cite{World_Models}). Especially in \cite{Brunnbauer2021} it is shown that model-based RL approaches outperform non-model based methods. In addition, the authors display the sim-to-real transfer by testing the trained agent on a F1TENTH vehicle. The vehicle shows good generalization on unknown tracks but no high performance (e.g. low laptime) because of oscillating steering.

\subsection{Applied Autonomous Racing Studies}
\label{subsec_additioalsoftware}

In the final subsection all applied autonomous racing studies are displayed that clearly do not belong in the previous sections \ref{subsec_perception}-\ref{subsec_end2end}. These research papers provide \textit{Evaluations} that run either simulator studies or overall analysis in the field of autonomous racing. In addition, the efforts of \textit{Complete Software Stack} developments are displayed here. In order to achieve the vehicle's driving dynamics limits, there must be in-depth knowledge of the driving dynamics behaviour and thus sufficiently good vehicle dynamic modeling. In the category of  \textit{Modelling} the research that shows all vehicle dynamic modelling efforts for later usage in either trajectory planner or control algorithms is displayed. Finally, we present \textit{Simulation} efforts and environments for autonomous racecars. A summary and overview of the research in those categories can be found in Table \ref{tab:software_additionalsoftware}.

\begin{table*}[htbp]
\centering
\caption{Overview and categorization of applied research and development studies in the field of autonomous racing}
\label{tab:software_additionalsoftware}
\begin{tabular}{|>{\raggedleft\arraybackslash}m{3.1cm}|>{\centering\arraybackslash}m{0.5cm}|>{\centering\arraybackslash}m{3.0cm}|>{\centering\arraybackslash}m{3.2cm}|>{\centering\arraybackslash}m{1.4cm}|>{\centering\arraybackslash}m{2.1cm}|}
\hline
\multicolumn{1}{|c|}{\textbf{Name and Reference}} & \multicolumn{1}{c|}{\textbf{Year}} & \multicolumn{1}{c|}{\textbf{Applied Studies Category}} & \multicolumn{1}{c|}{\textbf{Topic/Methods}} & \multicolumn{1}{c|}{\textbf{\begin{tabular}[c]{@{}c@{}}Tested on\\ Hardware\end{tabular}}} & \multicolumn{1}{c|}{\textbf{\begin{tabular}[c]{@{}c@{}}Racing \\ Series\end{tabular}}} \\ \hline
Samper et al.~\cite{SamperMejia2013}            & 2014 & Evaluation              & Path Analyzation           & Yes      & -   \\ \hline
Kegelman et al.~\cite{Kegelman2016}             & 2016 & Evaluation              & Simulator Study            & Yes      & -   \\ \hline
Remonda et al.~\cite{Remonda2021}               & 2021 & Evaluation              & Simulator Study            & No       & - \\ \hline
Betz et al.~\cite{Betz2019_3}                   & 2019 & Evaluation              & Crash Analysis             & -        & Roborace   \\ \hline
Hermansdorfer et al.~\cite{Hermansdorfer2020}   & 2020 & Evaluation              & Race Driver vs. Car                 & -        & Roborace   \\ \hline
Bak et al.~\cite{Bak2021}                       & 2020 & Evaluation              & Path Planner Stress Tester & No        & F1TENTH   \\ \hline
Funke  et al.~\cite{Funke2012}                  & 2019 & Complete Software Stack & Framework                  & Yes      & Racecar   \\ \hline
Culley  et al.~\cite{Culley2020}                & 2019 & Complete Software Stack & Framework                  & Yes      & FSD   \\ \hline
Funk  et al.~\cite{funk2017}                    & 2017 & Complete Software Stack & Framework                  & Yes      & FSD   \\ \hline
Tian  et al.~\cite{tian2018}                    & 2018 & Complete Software Stack & Framework                  & Yes      & FSD   \\ \hline
Chen  et al.~\cite{chen2019}                    & 2019 & Complete Software Stack & Framework                  & Yes      & FSD   \\ \hline
Zadok  et al.~\cite{Zadok2019}                  & 2019 & Complete Software Stack & Simulation                 & No       & FSD   \\ \hline
Kabzan  et al.~\cite{Kabzan2019}                & 2020 & Complete Software Stack & Framework                  & Yes      & FSD   \\ \hline
Nekkah  et al.~\cite{nekkah2020}                & 2020 & Complete Software Stack & Framework                  & Yes      & FSD   \\ \hline
Tian  et al.~\cite{tian2020}                    & 2020 & Complete Software Stack & Framework                  & Yes      & FSD   \\ \hline
Betz  et al.~\cite{Betz2019,Betz2019_2}         & 2019 & Complete Software Stack & Framework                  & Yes      & Roborace   \\ \hline
Caporale  et al.~\cite{Caporale2019}            & 2019 & Complete Software Stack & Framework                  & Yes      & Roborace   \\ \hline
You  et al.~\cite{You2017}                      & 2017 & Modelling               & UKF                        & Yes      & AutoRally   \\ \hline
Williams  et al.~\cite{williams2019}            & 2019 & Modelling               & Deep Learning              & Yes      & AutoRally   \\ \hline
Park  et al.~\cite{Park2017}                    & 2017 & Modelling               & Modelling                  & Yes      & -   \\ \hline
Spielberg  et al.~\cite{Spielberg2019}          & 2019 & Modelling               & Deep Learning              & Yes      & -   \\ \hline
Ignat  et al.~\cite{Ignat2020}                  & 2020 & Modelling               & Deep Learning              & Yes      & Roborace   \\ \hline
Hermansdorfer  et al.~\cite{Hermansdorfer2021}  & 2021 & Modelling               & Deep Learning              & Yes      & Roborace   \\ \hline
Wymann et al.~\cite{Torcs2005}                  & 2005 & Simulation              & TORCS Simulator            & No       & -   \\ \hline
Jiang  et al.~\cite{Jiang2021}                  & 2021 & Simulation              & Carla Simulator            & No       & -   \\ \hline
Guodong  et al.~\cite{Guodong2020}              & 2020 & Simulation              & SVL Simulator              & Yes      & F1TENTH, IAC   \\ \hline
Babu  et al.~\cite{Babu2020}                    & 2020 & Simulation              & F1TENTH Simulator          & Yes      & F1TENTH   \\ \hline
Stahl  et al.~\cite{Stahl2020_2}                & 2020 & Simulation              & Scenario Creation          & -        & Roborace   \\ \hline
Herman  et al.\cite{herman2021}                 & 2021 & Simulation              & Roborace Simulator         & Yes      & Roborace   \\ \hline
\end{tabular}
\end{table*}

\begin{figure}[h]
\begin{center}
\includegraphics[scale=0.23]{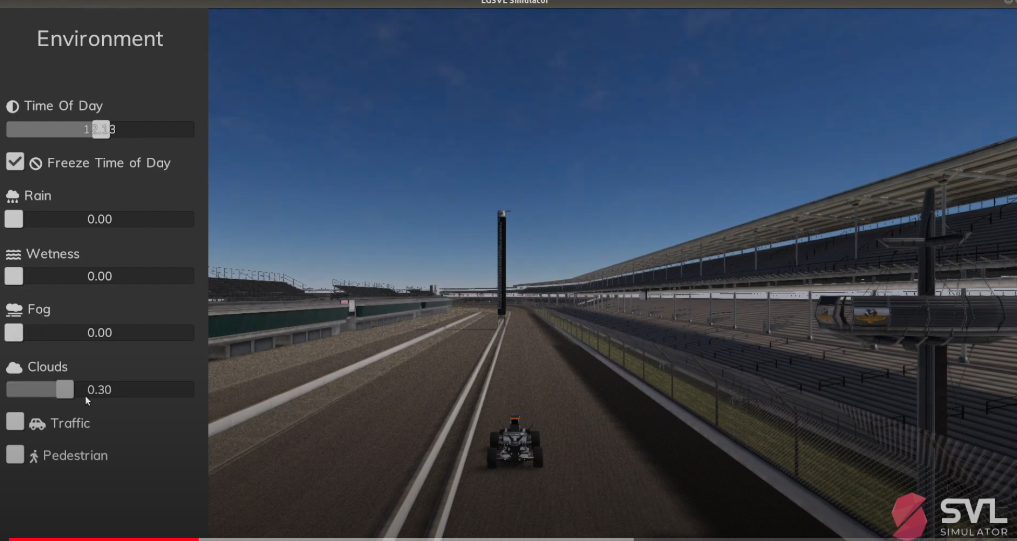}
\caption{F1TENTH vehicle in the SVL simulator \cite{Guodong2020} on a 1:10 scale version of the Indianapolis Motor Speedway.}
\label{fig_additionalsoftware_SVL}
\end{center}
\end{figure}

\noindent\textbf{Evaluations}\\
\indent To gain more knowledge in the field of racing different researchers conducted studies with race drivers or racecars. Kegelmann et al.~\cite{Kegelman2016} did a study with real (vintage) racing cars and collected vehicle and position data from their runs on the racetracks. By examining the statistical dispersion of the vehicles race lines, the author displayed a quantification of the the repeatability of professional racecar driver performance. In addition, they concluded that different driving styles (combination of path and velocity) can lead to similar lap times.  In \cite{SamperMejia2013} the raceline trajectory information (dGPS data) from a test vehicle is collected to derive a path fitting algorithm that is describing a raceline. Based on this setup the race lines can be analyzed in-depth and results for autonomous raceline planning can be derived. A direct comparison between autonomous racecar against a human race driver is both done in \cite{Remonda2021} and \cite{Hermansdorfer2020}. Remonda et al.~\cite{Remonda2021} conduct this study in a simulator environment and compared the lap times, telemetry data and the performance level of a human race driver against a pure autonomous racing software based on RL. By doing this the researchers were able to analyze  which features have the most impact on the drivers performance. Those features where used afterwards to enhance the RL approach. In contrast, Hermansdorfer et al.~\cite{Hermansdorfer2020} conduct a real world study by comparing an autonomous racing stack on the Roborace vehicle against a professional Formula 2 driver to find indications where the autonomous car fails to meet the performance level of the human race driver. The main reasons are that the human driver is detecting the vehicle limits (tire limits) more accurate, bringing the vehicle more often beyond the limit (higher slip angle) and is applying both brakes later and throttle earlier. Finally in \cite{Betz2019_3} an evaluation about a crash of an autonomous racecar is displayed. \\
\textbf{Complete Software Stack}\\
\indent Although many researchers are just deploying a single algorithm for testing and evaluation, to fully run an autonomous vehicle a complete software stack consisting of perception, planning and control algorithms is necessary. Therefore, many publications aim towards designing a holistic autonomous software stack that describes the individual software components, the methods, the transfer of messages from one module to the other and a final evaluation on real hardware or simulation. In \cite{Culley2020,Agnihotri2020,tian2020,funk2017,chen2019,Zadok2019,Kabzan2019,nekkah2020} the efforts of developing an autonomous racing software stack for FSD vehicles are presented. Based on the tasks in the FSD competitions (Section \ref{sec_hardware}) these cars need to map the environment, localize themselves, plan the path on the fly and follow the path fast and reliable. The teams provide different concepts to solve those individual tasks and display at the same time the underlying computation hardware of the their autonomous race vehicles. In addition, the teams provide insights in the middleware (e.g. ROS) as well as computations times of their algorithms. In contrast to the FSD efforts the publications \cite{Betz2019, Betz2019_2, Caporale2019} show their autonomous racing software stacks for running the Roborace vehicle. In \cite{Betz2019_2} the research is aiming towards a software that can operate in a  multi-vehicle scenario and therefore displays a dynamic local trajectory planner as a main component. In addition, to achieve high dynamic trajectory planning maneuvers the team displays a \textit{Performance Assessment Module} that is observing the controller and the tires while adjusting parameters accordingly. Caporale et al.~\cite{Caporale2019} display a holistic planning and control stack that has a real-time NMPC as main backbone to track a pre-planned racing line as well as a mapping and localization approach for high speed driving. \\
\textbf{Modelling}\\
\indent The modelling of the vehicle dynamics behavior of the racecar is an essential part in the field of autonomous racing. Either these models are used in the simulation environments or model-based trajectory planning/control design approaches. The current state of the art provides many variations of vehicle dynamics modeling such as single track model, double track model or full vehicle model. The more complicated the vehicle dynamics model, the more parameters are needed. Unfortunately not all of those parameters are available in detail for a vehicle and so different methods for estimating these parameters are proposed - especially for nonlinear vehicle parameters like the tires. In \cite{You2017} the design of the standard joint-state Unscented Kalman Filter (UKF) is presented to estimated vehicle dynamics parameters of a model car both in simulation and with experimental data. The experimental results show satisfactory estimates of the model parameters. Unfortunately the tuning process of this algorithm is time consuming and can only be implemented offline. Park et al.~\cite{Park2017} describe the region of feasible tire forces mathematically with constraints on the limits of actuation. They conclude in their work that with reasonable assumptions, the border of feasible tire forces can be displayed by an ellipse and circle for a wheel with steering and braking actuators. The papers \cite{williams2019, Spielberg2019,Ignat2020,Hermansdorfer2021} are using learning based approaches by applying DNNs to identify the model parameters. All of these works show that DNNs can learn and identify the vehicle parameters more accurately than a purely parametric model. In addition, the researchers showed that the DNNs can generalize better than a purely non-parametric model especially when it comes to capturing the unknown dynamics. This makes the usage of DNNs ideal for real-world applications where collecting data from the full state space for a vehicle is not feasible and when different environment dynamics (e.g. icy road) need to be captured in the model. \\
\textbf{Simulation}\\
\indent The final paragraph in this subsection displays all simulation efforts that have been done in the field of autonomous racing. Wymann et al.~\cite{Torcs2005} are the authors of \textit{TORCS - The Open Racing Car Simulator}. This lightweight 3D simulator provides different race tracks, different cars, NPC opponents as well as a sophisticated vehicle physics model. This simulator is used for research in the field of control, trajectory planning, game theory and RL and is therefore providing a solution for autonomous race engineers. Roborace released its own simulation environment that is enhanced with an openAI Gym interface especially for classical control or RL tasks \cite{herman2021} which they called \textit{Learn-to-Race (L2R)}. The simulation environment provides a racetrack, sophisticated vehicle physics simulation as well as a wide variety of sensors. The SVL Simulator \cite{Guodong2020} is a 3D end-to-end autonomous vehicle simulation platform that provides different maps, vehicles, sensor modelling, weather simulation, APIs to well-known open source software stacks (e.g. Autoware.Auto, Baidu Apollo) and the possibility of a distributed simulation. The SVL Simulator is offering both a 3D-vehicle model of the F1TENTH and the Indy Autonomous Challenge vehicle with different racetracks (Figure \ref{fig_additionalsoftware_SVL}).

For the F1TENTH vehicle different simulation environment exist. Babu et al.~\cite{Babu2020} present a ROS and Gazebo based autonomous racing simulator that is providing different maps, visualizations and model physics. The advantage here is access to the ROS community that enables the integration of robotics packages. Another F1TENTH simulator is the F1TENTH Gym \cite{okelly2019_2} that provides a lightweight, 2D-simulation with an openAI~Gym interface. Based upon the Carla Simulator \cite{Carla17} the authors of \cite{ICRAworkshop_13} present an Autonomous System Operations (AutOps) and continuous integration (CI) and testing framework to evaluate the software in the context of autonomous racing.  Especially for the evaluation of trajectory planning maneuvers in a multi vehicle environment Stahl et al.~\cite{Stahl2020_2} display an open-source graphical user interface that allows the fast generation of multi vehicle race scenarios. These dedicated scenarios (e.g. overtaking maneuvers) can be used to evaluate the trajectory planner or safety assessment algorithms in an autonomous racing stack. 

\section{Autonomous Racing Hardware: Vehicles and Competitions}
\label{sec_hardware}
The previous section gave an overview on the efforts in the field of algorithm and software development for autonomous racing vehicles. Almost all of the papers provided an evaluation of their proposed methods in an specific simulation environment. About half of the papers did additional evaluations on real vehicle hardware. This hardware is ranging from (powerful) passenger sports cars, specific research vehicle prototypes, small-scale race vehicles or real racing cars. In the following section we provide an overview of currently available hardware and racing competitions (Table \ref{tab_hardware_overview}) that are available for researchers.

\begin{table*}[h!]
\centering
\caption{Autonomous Racing Hardware: Overview over different available hardware and racing competitions available for researchers}
\label{tab_hardware_overview}
\begin{tabular}{|>{\centering\arraybackslash}m{1.7cm}|>{\centering\arraybackslash}m{2.7cm}|>{\centering\arraybackslash}m{2.7cm}|>{\centering\arraybackslash}m{2.7cm}|>{\centering\arraybackslash}m{2.7cm}|>{\centering\arraybackslash}m{2.7cm}|}
\hline
\multicolumn{1}{|l|}{\textbf{}} & \textbf{F1TENTH\footnotemark[3]} & \textbf{EV Grand Prix Autonomous\footnotemark[4]} & \textbf{Formula Student Driverless\footnotemark[5]} & \textbf{Indy Autonomous Challenge\footnotemark[6]} & \textbf{Roborace\footnotemark[7]} \\ \hline
\textbf{Vehicle Image} & \includegraphics[width=0.120\textwidth]{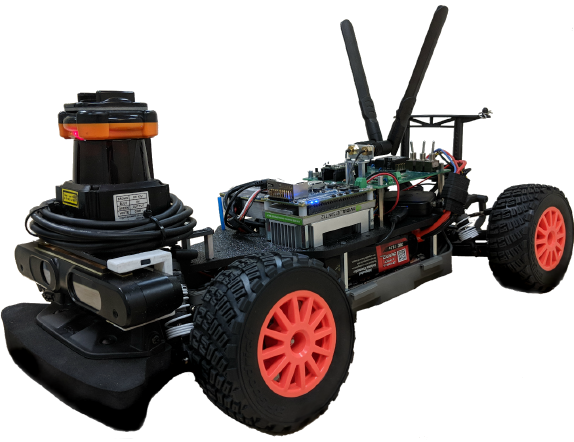} & \includegraphics[width=0.125\textwidth]{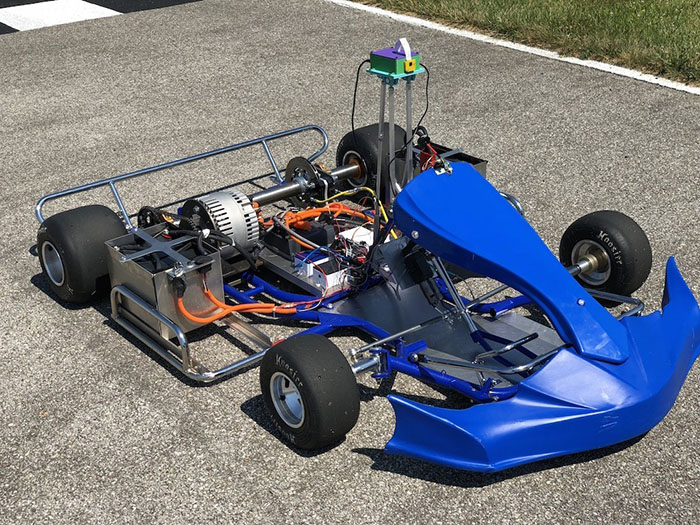} & \includegraphics[width=0.126\textwidth]{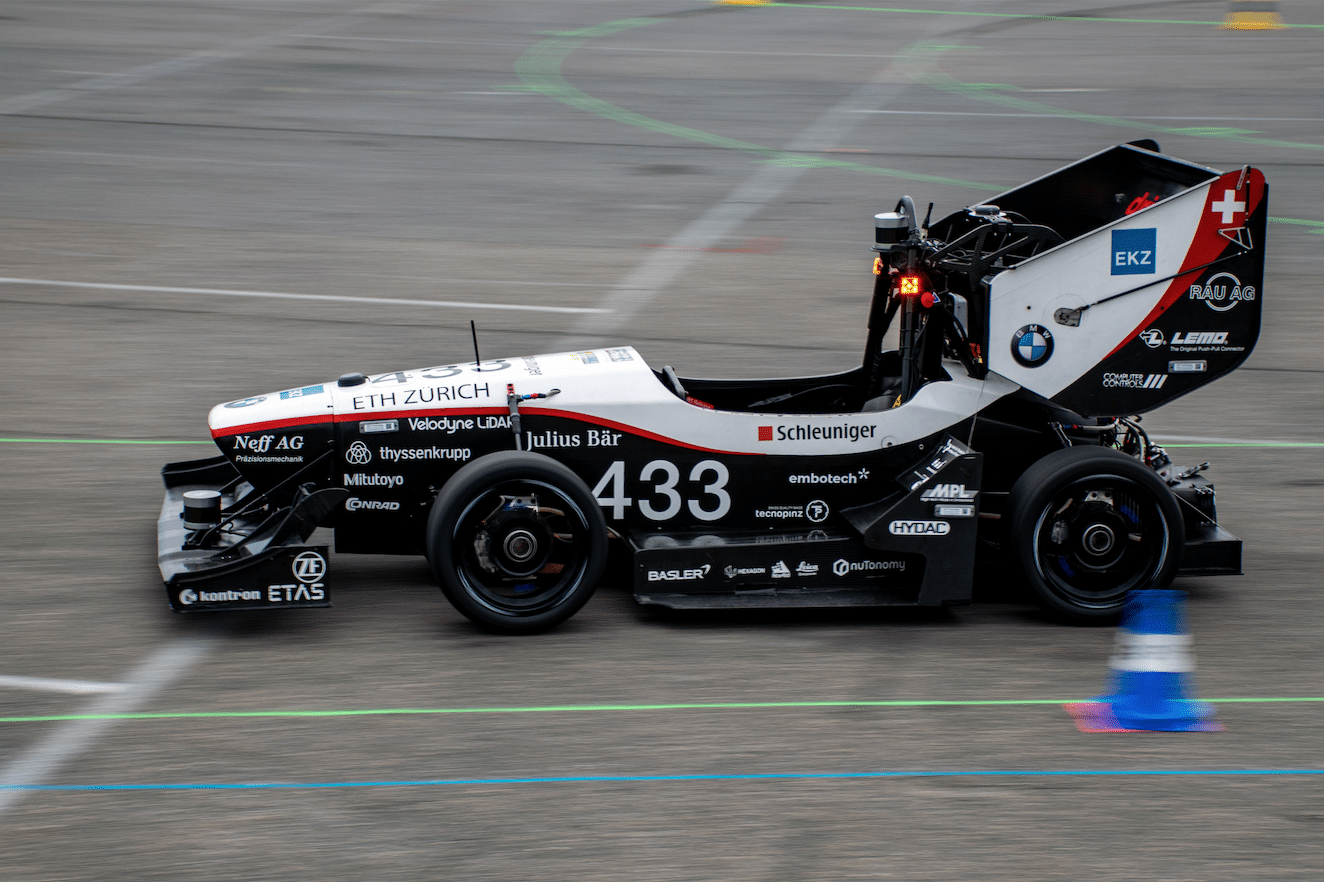} & \includegraphics[width=0.126\textwidth]{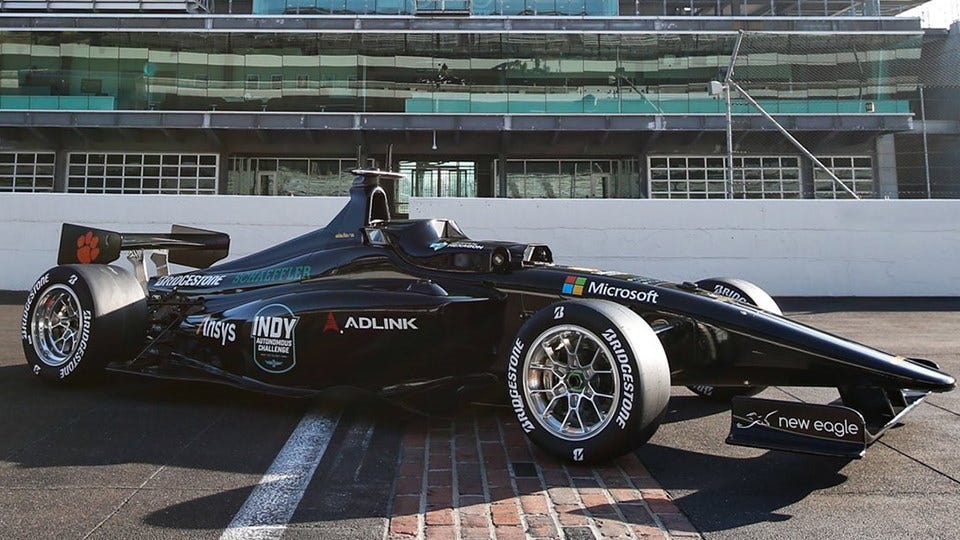} & \includegraphics[width=0.126\textwidth]{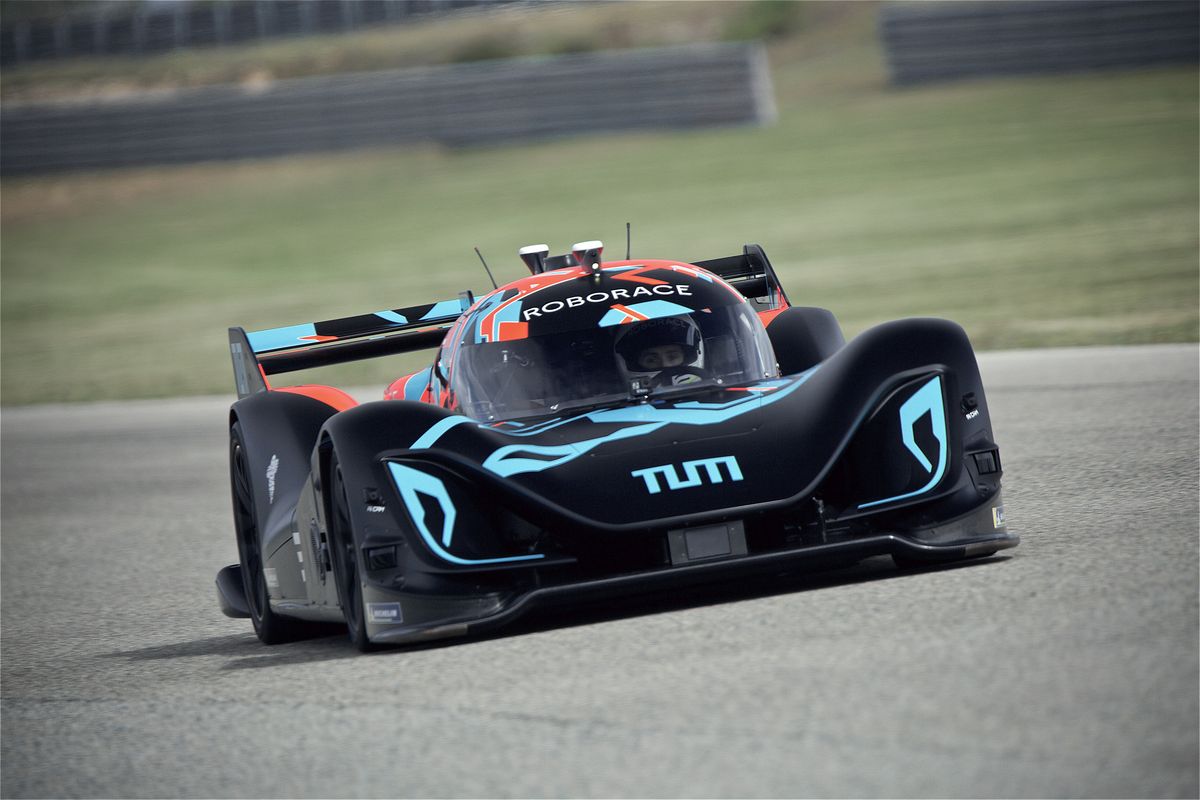} \\ \hline

\textbf{Vehicle Type}  & \makecell{Small Scale\\ 1:10} & \makecell{Reduced Scale\\ 1:3}  & \makecell{Reduced Scale\\ 1:1.5}  & \makecell{Real Racecar\\ Indy Light Chassis}  & \makecell{Real Racecar\\ LMP Chassis}  \\ \hline

\textbf{Vehicle Parameters} & \makecell{Length: 0.53~m\\ Width: 0.28~m \\ Mass: 3.5~kg }  & \makecell{Length: 1.5~m\\ Width: 1.4~m \\ Mass: 110~kg } & \makecell{Vehicle parameters are \\ based on the teams \\ design choices} & \makecell{Length: 4.9~m\\ Width: 1.9~m \\ Mass: 750~kg } & \makecell{Length: 4.7~m\\ Width: 2.0~m \\ Mass: 1200~kg } \\ \hline

\textbf{Powertrain} & \makecell{Electrical Engine\\ AWD \\ Engine: 230~W \\ Battery: 55~Wh } & \makecell{Electrical Engine\\ RWD \\ Engine: Teams choice \\ Battery: Teams choice}  & \makecell{Electrical Engine\\ AWD/RWD \\ Engine: Teams choice \\ Battery: Teams choice}  & \makecell{Combustion Engine\\ RWD \\ Engine: 335 kW  \\ 6 speed sequential } & \makecell{Electrical Engine\\ RWD \\ Engine: 270 kW  \\ Battery: 40 kWh }  \\ \hline

\textbf{Maximum Speed} & $\sim$72 km/h & \makecell{$\sim$ 100 km/h \\ Depends on  \\components choices}  & \makecell{$\sim$ 120 km/h \\ Depends on  \\components choices} & $\sim$290 km/h & $\sim$250 km/h  \\ \hline

\textbf{Sensor Setup} & \makecell{Monocular camera\\ Stereocamera \\ 2D LiDAR \\ Indoor GPS} & \makecell{Sensor setup based \\ on team choice: \\ Monocular camera\\ Stereocamera \\ Radar\\ 2D LiDAR\\3D LiDAR \\ (RTK) GPS} & \makecell{Sensor setup based \\ on team choice: \\ Monocular camera\\ Stereocamera \\ Radar\\ 2D LiDAR\\ 3D LiDAR \\ (RTK) GPS} & \makecell{6x Monocular camera\\4x Radar \\ 3x 3D LiDAR \\ (RTK) GPS} & \makecell{4x Monocular camera\\2x Long Range Radar\\2x Short Range Radar \\ 5x 3D LiDAR \\ (RTK) GPS}  \\ \hline

\textbf{Computation Unit} & \makecell{Nvidia Jetson Nano\\ Nvidia Jetson NX\\ Nvidia Jetson  AGX}  & Teams choice & Teams choice & Intel Xeon E 2278 GE – 3.30 GHz, 1x Nvidia Quadro RTX 8000, 64GB Ram  & \makecell{Nvidia Drive PX2\\ Speedgoat Mobile\\ McLaren ECU} \\ \hline

\textbf{Software} & \makecell{ROS\\ ROS2 \\ Autoware.Auto}  & Teams choice  & Teams choice & \makecell{ROS2 \\ Autoware.Auto} & Teams choice\\  \hline
\textbf{Competitons} & Several competitions a year, competitions in different countries & One Race, USA only & Several competitions a year, competitions in different countries & Two Races, USA only &  One championship with several races, USA \& UK\\ \hline
\textbf{Single/Multi Vehicle Race} & \makecell{Multi Vehicle \\ 2 Cars} & Single Vehicle & Single Vehicle & \makecell{Single Vehicle \\ Multi Vehicle (2 Cars)} & \makecell{Single Vehicle \\ Multi Vehicle (2 Cars)}  \\ \hline

\textbf{Real Race Competition Type} & \makecell{Time Trial \\Head to Head Racing} & \makecell{Time Trials}  & \makecell{Acceleration \\ Skidpad \\ Autocross \\ Trackdrive \\ Efficiency \\ Business, Cost, Design} & \makecell{Time Trial\\ Overtaking Competition} & \makecell{Time Trial\\ (with virtual objects)}\\ \hline
\textbf{Virtual Race} & Yes & No & No & Yes & No  \\ \hline

\textbf{Simulation Environments} & \makecell{F1TENTH Gym \\ F1TENTH Simulator \\ SVL Simulator}  & - & \makecell{Formula Student \\ Driverless Simulator}   & \makecell{Ansys Simulator\\ SVL Simulator}   & Roborace Simulator  \\ \hline

\textbf{Related Papers} & \cite{okelly2019,okelly2019_2,OKelly2020,Ivanov2020,Wang2021,Babu2020,Agnihotri2020,evans2021,evans2021_2,Joglekar2021, sinha2020, Bulsara2020,jain2020,Tatulea2020,Brunnbauer2019,Brunnbauer2021, Gotlib2019,ICRAworkshop_07}  & - & \cite{Kabzan2019,Kabzan2020,Liu2020,Dhall2019,Ji2018,Andresen2020,srinivasan2021,Vazquez2020, DeRita2019, Puchtler2020, Valls2018,Zeilinger2017, tian2018, tian2020, chen2019, Zadok2019, nekkah2020, funk2017, Ni2017, Ni2019, dodel2021, Strobel2020,Gosala2019, Feraco2020, LeLarge2021, Tramacere2021} & \cite{ICRAworkshop_01, ICRAworkshop_06,ICRAworkshop_10} & \cite{Hermansdorfer2019,Hermansdorfer2020,Hermansdorfer2021, Betz2018,Betz2019, Betz2019_2,Betz2019_3,Caporale2018,Caporale2019, Zubov2018, Herrmann2019,Herrmann2020,Herrmann2020_2, Chatzikomis2018, Heilmeier2019,Zubaca2020, Zubaca2020_2, Wischnewski2019, Wischnewski2019_2, Wischnewski2020, Wischnewski2021, Stahl2019, Stahl2019_2, Stahl2020, Stahl2020_2, Stahl2021, Massa2020, Renzler2020,ICRAworkshop_04, Christ2019,Ignat2020, Buyval2017, Nobis2019, Palafox2019, herman2021} \\ \hline
\end{tabular}
\end{table*}

\subsection{Small-Scale Autonomous Racing Vehicles}
\label{subsec_smallscale}

\indent The first type of autonomous racing vehicles are so called \emph{small-scale} or \emph{reduced-scale} vehicles. These racecars were mainly developed for the purpose of testing the new developed autonomous racing software. Those racecars are normally derived from remote controlled (RC) cars and therefore provide an electrical engine and a battery as a main powertrain unit. Those vehicles are then modified with additional hardware (sensors, ECUs), are constructed and maintained by a team of students and researchers and usually costs a few hundred to a few thousands of dollars. Although these are small-scale vehicles, they reach high speeds and accelerations for their size and therefore can be compared to real racecars.
\footnotetext[1]{https://control.ee.ethz.ch/research/team-projects/autonomous-rc-car-racing.html}

\noindent \textbf{1:43 vehicles}\\
\indent In the ORCA (Optimal RC Racing) \footnotemark[1] project researchers from the ETH Zurich developed a test bed consisting of a race track, an infrared camera based tracking system and modified 1:43 cars, in order to apply research in the field of MPC algorithms at high speeds and in real-time. A vision system captures the cars on the track and estimates both positions and velocity of each car. This information is then sent to a specific control platform where the MPC controller calculates the control inputs for the cars. This information is then sent via Bluetooth to the embedded cars where the control input is actuated. The research published with this 1:43 cars is heavily in the field of planning and control. As a result the researchers displayed new developments in the field of MPC \cite{Liniger2014,Liniger2015, Liniger2017, Liniger2018, Hewing2018, Carrau2016}, game theory \cite{Liniger2020} and reinforcement learning \cite{chisari2021}.  \\ \\

\noindent \textbf{1:10 vehicles}\\
\indent  In the next bigger size researchers use modified 1:10 scale RC cars for their autonomous racing research. In the last years different institutions released their documentation for both hardware and setup of these 1:10 vehicles and so currently versions like \textit{the Berkeley Autonomous Racecar}\footnotemark[2], the \textit{MIT Racecar} \cite{Karaman2017}, the \textit{MuSHR racecar} \cite{srinivasa2019}, the \textit{RoSCAR} \cite{Hart2014} or the \textit{F1TENTH} \cite{okelly2019,okelly2019_2} vehicle  from from the University of Pennsylvania exists. The sensor setup on these cars is interchangeable and so it is possible to apply monocular cameras (e.g. Raspberry Pi, OpenCV OAK-1), stereo cameras (e.g. ZED, ZED2, Intel Realsense d435i, OpenCV OAK-D), 2D LiDARs (Hokuyo 10LX, Hokoyu 20LX), IMU, indoor GPS or wheelspeed sensors. As a main computation platform these vehicles use embedded GPU systems like the Nvidia Jetson (Models: TX1, TX2, NX, AGX Xavier, Nano). This gives the possibility to speed up the inference of DNNs. With the F1TENTH vehicle an additional, annual autonomous racing competition was launched where students, researchers and hobbyists can race against each other. The competitions consists of a single vehicle time trial and a head-to-head two vehicle race with knockout phase. In addition to these in-person competitions virtual competitions are organized to test the software of the developers. Similar competitions, where those kind of racecars or variations of it are used, are the \emph{DiYRobocar} events or the Amazon \emph{DeepRacer} \cite{balaji2019,Balaji2020} competitions. 
The research published with these 1:10 cars is spread completely over all topics in perception \cite{Brunnbauer2019,Gotlib2019}, planning \cite{Joglekar2021,ICRAworkshop_07} and control \cite{Bulsara2020,Ivanov2020,jain2020, Tatulea2020,ICRAworkshop_12, Rosolia2020, Brunner2017, Pagot2020}. In recent years these type of vehicles got more important for optimization pipelines \cite{OKelly2020,sinha2020}, the application of RL techniques \cite{Brunnbauer2021,evans2021, evans2021_2} and the evaluation of game theory methods \cite{Wang2019,Wang2021}. In addition,  the 1:10 vehicles are used for education purposes \cite{Agnihotri2020,Babu2020,Karaman2017,Eken2020} to teach students hands-on fundamentals of autonomous driving. 

\footnotetext[2]{www.barc-project.com/projects}
\noindent \textbf{1:5 vehicles}\\
\indent A special version of an autonomous small-scale vehicle is the so called \emph{AutoRally} \cite{Goldfain2019} vehicle, a 1:5 scale autonomous racecar developed by a team of researchers from Giorgia Tech. The AutoRally autonomous vehicle platform is based on a RC trophy truck (length: 1~m, width: 0.6~m, mass: 22~kg) with a top speed of $\sim$ 90 km/h. The AutoRally vehicle uses two monocular cameras (Point Grey Flea3 FL3-U3-13E4C-C color) as a main sensors to perceive the environment, has an IMU for acceleration measurements and hall-effect sensors to measure the wheel speeds. In addition, this vehicle has a GPS receiver (Hemisphere P307) integrated which provides an absolute position at 20 Hz with an accuracy of approximately 2~cm under ideal conditions with Real-Time Kinematic (RTK) corrections that are derived from a GPS base station. The main computation unit consists of standard consumer computer components (Intel i7-6700 -3.4 GHz quad-core, 32 GB DDR4, Nvidia GTX-750ti SC) which are modular and reconfigurable and are all connected on a Mini-ITX motherboard. 

\footnotetext[3]{www.f1tenth.org}
\footnotetext[4]{www.evgrandprix.org/autonomous/}
\footnotetext[5]{www.fsaeonline.com}

\noindent This brings the AutoRally setup closer to real passenger vehicles and allows  a high computation power. The research published with the AutoRally car is in the field of planning \cite{Arslan2017, Lee2020, You2018, You2021, Williams2017} and control \cite{Drews2019,Gandhi2021,Lee2019, wagener2019, Williams2016, Williams2017_2, Williams2018, Williams2018_2, Williams2018_3, williams2019}, state estimation \cite{Foris2020,You2017} and the application of deep neural networks for perception and planning \cite{Drews2019,Pan2018} with an overall special focus on low friction surfaces.

\footnotetext[6]{www.indyautonomouschallenge.com}
\footnotetext[7]{www.roborace.com}

\noindent \textbf{eV Grand Prix Autonomous}\\
\indent In 2021 a new racing competition called \emph{eV Grand Prix Autonomous} started in the USA. Student teams need to acquire a standardized electric go-kart chassis and are then allowed to modify the vehicle. The teams can change the complete electric drivetrain and integrate new components, e.g. battery, electrical engine, inverter. In addition,  the teams can choose their own sensor setup (camera, radar, LiDAR, (RTK) GPS, IMU) to create an autonomous vehicle. Furthermore, the teams develop the software that drives the autonomous go-kart around the racetrack. The current race setup consists of single vehicle time trial laps. Because the eV Grand Prix autonomous is still in its early stage there was no research published with these kind of vehicles so far. \\
\noindent \textbf{Formula Student Driverless}\\
\indent Since 2017 student teams can develop a driverless vehicle for the Formula Student Driverless competition. The students can choose on their own how to design and equip the vehicle with both powertrain or autonomous driving hardware. Therefore different setups with different computation platforms (e.g. consumer hardware, Nvidia Drive PX hardware, Nvidia Jetson hardware) and sensors setups (monocular cameras, 3D LiDAR) exist. The cars compete in different single vehicle competitions: Acceleration (driving 75m straight with standing start), skid pad (two congruent circles with a diameter of 18.25m), autocross (racing on closed loop track with unknown layout), track drive and efficiency (racing 10 laps on a track with additional efficiency scoring based on the consumed energy). In addition to these pure driving competitions the cars are then judged in an additional business plan (business idea of the vehicle), design (judgement of hardware and software) and cost (financial planning an manufacturing) competition.

The research published with the Formula Student cars is spread completely over all topics in object detection \cite{dodel2021,DeRita2019,Dhall2019,Puchtler2020, Strobel2020}, localization \cite{Andresen2020,Gosala2019,LeLarge2021,Srinivasan2020}, planning \cite{Feraco2020, srinivasan2021, Vazquez2020} and control \cite{Fu2018,Ji2018, Kabzan2019,Liu2020, Ni2017,Ni2019} with a focus on holistic software pipelines \cite{Kabzan2020, chen2019, nekkah2020, tian2018, tian2020, Zadok2019, Zeilinger2017, Valls2018, Drage2014, funk2017} with adjustments for the specialities of the FSD competition.

\subsection{Real Autonomous Race Vehicles}
\label{subsec_realscale}

The small-scale vehicles offer a low-cost and easy to set up platform for researchers. In addition, only a small space is needed to run the vehicles and therefore these kind of small scale vehicles are very attractive for research in the field of autonomous racing. Unfortunately because of the scaling there is still a mismatch between those vehicles and real racecars. This has not only to do with the performance ($v_{max}$, $a_{long,max}$, $a_{lat,max}$) but also with the kind of sensors or computation units these vehicles equipped with. Furthermore, a real racecar has a different dynamic behavior because of the stiffness of the chassis. Based on this some companies/institutions decided do develop real autonomous race vehicles which are explored in further detail. There was an additional development of an autonomous dragster \cite{Bell2020,Bell2020_2}, the application of which, apart from in these papers, has not taken place elsewhere and is therefore not be considered in the further discussion. \\
\noindent \textbf{Roborace}\\
\indent Roborace is a UK based company that developed different autonomous racecars (Devbot 1.0, Devbot 2.0, Robocar). The Robocar was only used for internal Roborace events and both Devbot 1.0 and 2.0 where provided to interested university teams and companies for their research. The Devbot 2.0 is based on a Le Mans Prototype (LMP) chassis and is a rear wheel drive, fully electric racecar. The vehicle is equipped with camera, LiDAR and radar sensors and two main ECUs (Nvidia PX2, Speedgoat Mobile Target Machine) to run the autonomous software. The goal of this vehicle development efforts is to provide both a vehicle platform and an annual competition where teams can compete against each other. The teams only need to develop the software, the hardware setup is equal for all teams. In 2018 single vehicle time trials were executed, in 2019 the so called \emph{Season Alpha}  provided different race formats (single vehicle, multi vehicle, localization) on racetracks in Europe. In 2020/2021 Roborace \emph{Season Beta} started with seven university teams competing against each other in single-vehicle races (time trials). A special software from Roborace called \emph{Metaverse} provides virtual static and dynamic objects on the track that needed to be avoided by the teams - otherwise they get time penalties for hitting these objects.
The university teams used the Roborace vehicles for their research and provided plenty of published papers in the field of localization and motion estimation \cite{Massa2020, Renzler2020, Stahl2019, Wischnewski2019, Zubaca2020,ICRAworkshop_04}, mapping \cite{Nobis2019,Palafox2019}, planning \cite{Herrmann2020,Herrmann2020_2, Stahl2019, Stahl2020, Stahl2021, Heilmeier2019, Hermansdorfer2019, Christ2019, Caporale2018} and control \cite{Buyval2017, Chatzikomis2018, Wischnewski2019_2,Wischnewski2020, Wischnewski2021, Zubov2018} as well as energy management \cite{Herrmann2019,Herrmann2020} and holistic software stack development \cite{Betz2018,Betz2019,Betz2019_2, Betz2019_3}. In addition, the Roborace vehicle was used to derive some new simulation \cite{herman2021, Heilmeier2018, Heilmeier2019_2} and scenario environments \cite{Stahl2020_2}, vehicle dynamics modelling \cite{Hermansdorfer2021, Ignat2020} and autonomous racing benchmarks \cite{Hermansdorfer2020}.\\
\noindent \textbf{Indy Autonomous Challenge}\\
\indent In 2020 the \emph{Indy Autonomous Challenge} (IAC) was launched as a successor of the DARPA Grand Challenge and DARPA Urban Challenge. The IAC racecar is based on an Indy Lights chassis and is a rear wheel drive racecar powered by a combustion engine with 6 gear sequential transmission. The IAC vehicle is equipped with camera, LiDAR and radar sensors for perception and has one main ECU to run the autonomous software. The IAC provides universities both the vehicle platform and several competition types. The teams only need to develop the software, the hardware setup is equal for all teams. As a main middleware ROS2 is used. The IAC challenge consists of a single vehicle race around the Indianapolis Motor Speedway in October 2021 and a two vehicle head-to-head race on the Las Vegas Motor Speedway in January 2022. The aim is to drive 290 km/h with those vehicles and therefore the teams need to develop a high performance autonomous software stack that executes perception, planning and control. 
Since the IAC competition just finished its competition only a few papers \cite{ICRAworkshop_01,ICRAworkshop_06, ICRAworkshop_10} have been published so far.\\
\vspace{-0.5cm}

\section{Open Research Questions and Challenges}
\label{sec_openquestios}
In the previous sections we provided a detailed overview of all the efforts that have been made in the field of autonomous racing for both software and hardware. The goal of all this research efforts is to contribute to the development of safer autonomous passenger vehicles and the possibility to derive knowledge for the development of new and advanced autonomous driving algorithms. Although the state of the art is quite extensive, there are still some open and unsolved research questions where the field of autonomous racing can support, help and leverage the development of future autonomous driving algorithms. Based on additional discussions with leading researchers in the field we present a list of challenges in the field that determine open research questions:

\noindent \textbf{Challenge 1 - Autonomous high speed perception:} None of the previous work is covering high speed object detection or providing a detailed insight in different fusion techniques for high speed localization. The current state of the art presents standard SLAM or object detection methods that are then adapted to the field of autonomous racing. We are currently missing methods, techniques and algorithms that are especially made for high speed driving where increased motion blur occurs and sensor synchronization becomes more important. A reliable detection distance above 100~m is required. This can be achieved by decreasing the computational delay, an enhancement in the sensor fusion performance (camera+Radar+LiDAR) and with an increase of the object detection quality. When it comes to vision-based localization we see successful research in the field of drone racing which can be adapted and applied to the field of autonomous racing. Besides that there is currently no public dataset for high speed driving. However, the availability of rich data is essential for the development of comprehensive perception algorithms for this speed range.
    
\noindent \textbf{Challenge 2 - Multi vehicle trajectory planning:} Most of the displayed papers are dealing with a single vehicle setup and only a handful of researchers tried to address multi vehicle scenarios ($>$3 vehicles). Dynamic local trajectory planning at high speeds with multiple vehicles (e.g. for overtaking) is difficult and not covered completely in the state of the art and displays therefore a grand challenge for future research. The trajectory planning method must be capable of finding a path in a non-convex environment that is collision free, recursive feasible, incorporating dynamic vehicle constraints and is executable in real-time. Both the path and the velocity must be planned while taking the vehicle dynamics into account to leverage the current tire performance of the vehicle. We see this either as a chance for creating new types of methods and algorithms for trajectory planning or as a test environment for new heuristics that decrease the computational heavy calculations. 
    
\noindent \textbf{Challenge 3 - Multi vehicle interaction:} The interaction with other vehicles is an essential part of racing especially when it comes to head-to-head racing (e.g. overtaking, blocking). This interaction is covered with game theory approaches in some of the work but is not explored extensively. This interaction provides the need for new prediction algorithms that can deal with the high uncertainty of the opponents movements/behavior in the less structured environment of the racetrack. Hence the prediction of surrounding objects can not rely on lane information or traffic rules, but has to be based on a comprehensive understanding of interactive scenarios. Besides that a fundamental aspect of future state prediction is that it is inherently stochastic, as agents cannot know each other’s motivations, so multiple modalities have to be considered. We seek a model of the future that can provide both (1) a weighted, thrifty set of discrete trajectories that covers the space of likely outcomes and (2) a closed-form evaluation of the likelihood of any trajectory. In addition, there are no sophisticated behavior planners that can derive critical interaction based maneuvers for competing in a race environment. The goal is to enable a tight coupling between reason about the influence of the surrounding agents on the ego systems trajectory and maintaining full capabilities of the ego systems vehicle dynamics.
    
\noindent \textbf{Challenge 4 - Adversarial driving:} The racetrack enables the testing of the capabilities of an adversarial vehicle that is exploring and evaluating the risk of future actions by planning  and performing high risk maneuvers. This kind of research enables knowledge for autonomous cars that need to operate in highly crowded multi vehicle and multi passenger scenarios while minimizing the possibility of a freezing robot problem. This research heavily includes the calculation of risk for a perceived environment, a risk evaluation as well as new high precision local behavioral and trajectory planning algorithms. 
    
\noindent \textbf{Challenge 5 - Real-time vehicle dynamics modelling:} Autonomous vehicles that operate on the limits of handling need to have an exact knowledge about the current vehicle dynamics state. One crucial factor hereby is the tire which is creating the road-vehicle contact (friction value) which is changing drastically with aerodynamics (downforce), road conditions (tarmac), weather conditions (rain, snow) and the current vehicle maneuver (load shift due to braking, acceleration). The existing high model uncertainty due to external influences combined with strongly non-linear effects in tire and vehicle dynamics represents a challenge for both motion planning and control (e.g. MPC) algorithms. Available models approximate the vehicle dynamics to a certain degree but are computationally demanding, especially when it comes to tire models. Current research efforts try to calculate the dynamical behavior of the vehicle with the help of artificial neural networks to be computationally faster then classical physical models.
    
\noindent \textbf{Challenge 6 - Balancing safety and performance:} The current work is heavily exploring the limits of an autonomous vehicle from a software and hardware perspective with the goal to drive fast. When it comes to a racing scenario we have to make a trade off between safety (not crashing the car), high performance (staying close to the opponent), energy management and making decisions while not violating the handling limits (stay behind opponent, overtake in particular turn). This setup creates the need for software that explores the trade-off between safety and performance. This software is then coupled with motion and behavioral planners and decides which actions to take next. In addition, the current state oft the art does not cover the safety aspects of the autonomous racecar in particular and therefore we have an open research area where algorithms need to be derived that evaluate and balance both safety and performance of the vehicle.  
    
\noindent \textbf{Challenge 7 - Autonomous racing regulations and rulebook:} Although the community currently consists of many different racing series with different cars we have no agreement on the driving rules for autonomous racecars. Although this can be declared as less research and more a community effort, when it comes to autonomous driving a rulebook based definition for racecars could be helpful. This leads to general guidelines researchers can rely on when developing their algorithms. Ultimately, this leads to software that is compliant for specific racecar competitions.

\noindent \textbf{Challenge 8 – Overall software application:} Besides the development of particular algorithms for autonomous racing in each part of the software stack, the in-depth analysis of the overall software performance is a research field that is rarely covered. The synchronization of modules and the application of real-time conditions can reduce the overall latency significantly. Additionally, the delay of sensors and actuators influence both reaction time and vehicle performance. Unfortunately, these need to be considered in the development of a full software stack for autonomous racing. The optimal scheduling of software execution steps and an efficient management of the CPU and GPU usage by orchestration and hypervisor methods are additional research fields for the software development.

\noindent \textbf{Challenge 9 - Autonomous high speed hardware:} All platforms discussed in this survey are relying on standard consumer hardware (e.g. ECUs, sensors). There is currently no hardware existing that is aiming towards high speed driving or that was made particularly for autonomous racecars regarding computational demand as well as for vibration and shock resistance. Especially when we have a closer look to execution times displayed in the listed papers we see that some of the algorithms can be executed faster if particular hardware would exist.

\section{Summary and Conclusions}
\label{sec_summary}

\begin{figure}[h]
\begin{center}
\includegraphics[scale=0.45]{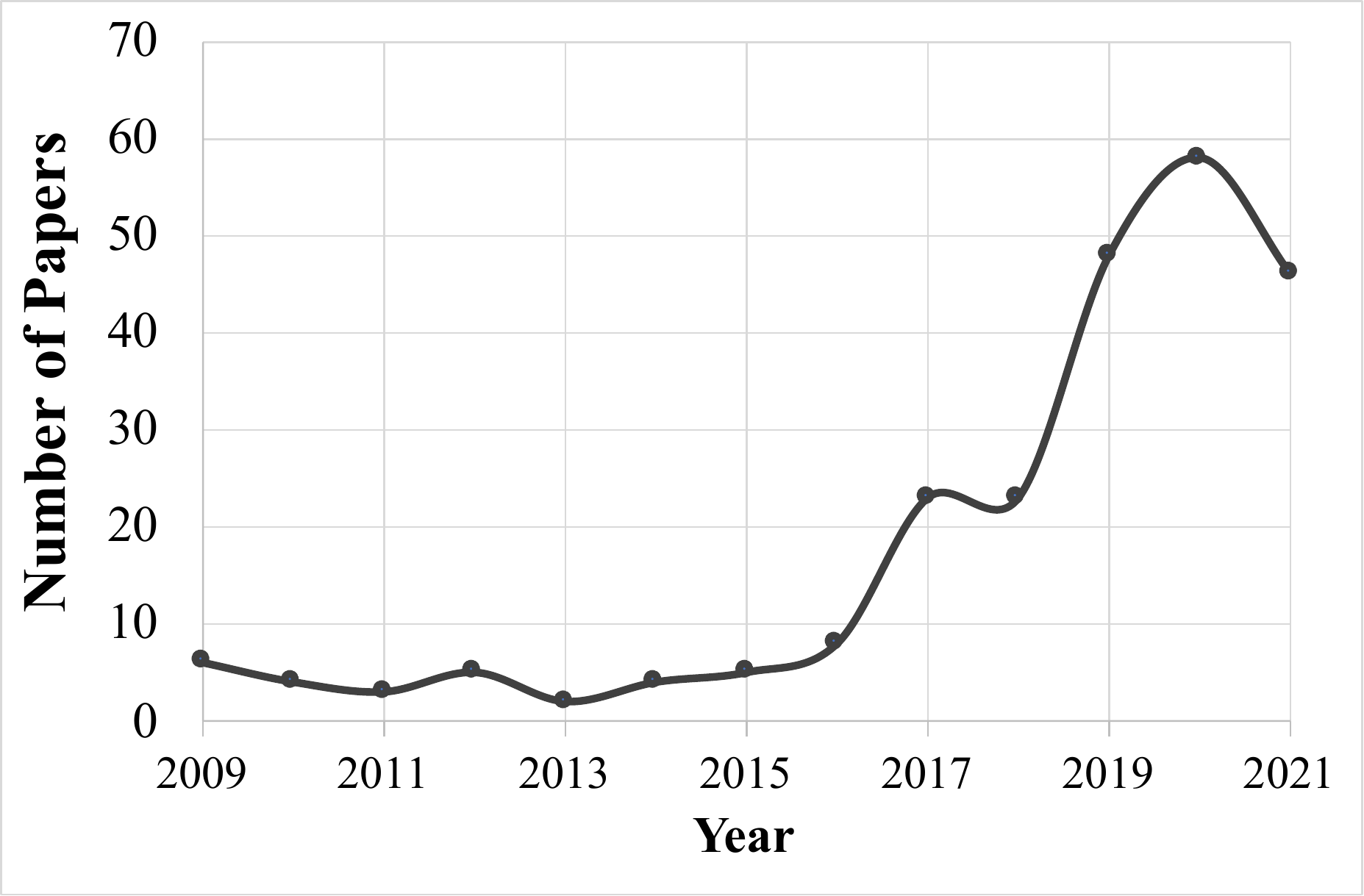}
\caption{Evaluation: Published paper in the field autonomous racing from 2009 until the end of 2021.}
\label{fig_summary_years}
\end{center}
\end{figure}

This survey paper presents a comprehensive overview of the current state of the art in the field of autonomous vehicle racing. By discussing the previous and ongoing research efforts in this field we were able to demonstrate what kind of algorithms were developed to derive autonomous high speed driving on the racetrack. By splitting this paper into different sections for perception, planning and control we showed the individual achievements derived by researchers to establish the autonomous driving task for a racecar. Furthermore, we displayed and categorized research in the field of end-to-end algorithms, vehicle dynamics modelling and simulation environments. This survey aims towards a holistic review in the field of autonomous racing so additionally all hardware developments and autonomous racing platforms that are available are explained in detail. Some of these vehicles are connected to regularly competitions that provide an additional platform for researchers to test and evaluate the performance of their software. In total this survey is covering 233 papers in the field of autonomous vehicle racing. Furthermore, in the last four years we saw an increase of papers in this field (Figure~ \ref{fig_summary_years}.

Undoubtedly, the field of autonomous racing is an emerging field in intelligent vehicles, robotics and transportation system that is attracting more and more interest from researchers. Although we see an overweight of research in the field of planning \& control, emerging fields like reinforcement learning are easily applicable to autonomous racing. Based on these results we derived a list of open research challenges for autonomous racing. This list  can be used as a guideline for future researchers which can participate in the displayed autonomous racing competitions.
Finally, the list of papers surveyed in this paper are uploaded to a  \href{https://github.com/JohannesBetz/AutonomousRacing_Literature}{Github repository} and updated on a regular basis so other researchers have an easy, open-source and structured access to the papers in the field of autonomous racing.

%\section{}

\section*{Contributions and Acknowledgement}
Johannes Betz initiated the idea of this paper, created the overall structure and contributed to all sections of this survey paper. Hongrui Zheng contributed to the path planning section. Alex Liniger and Ugo Rosolia contributed to the path planning and control section. Phillip Karle contributed to the path planning section and open research challenges. Madhur Behl and Venkat Krovi revised the paper critically. Rahul Mangharam contributed to the overall structure of the paper, the open research challenges and revised the paper critically. \\
%The research is partially funded by Carnegie Mellon University’s Mobility21 National University Transportation Center, which is sponsored by the US Department of Transportation and NSF CCRI: MEDIUM: Collaborative Research: Community Platforms for Safe, Agile, and Coordinated Autonomy (1925587). \\
\indent We would like to thank Todd Murphy (Northwestern University),  Davide Scaramuzza (University of Zurich), Chris  Gerdes (Stanford University), Markus Lienkamp (Technical University of Munich), Panagiotis Tsiotras (Georgia Institute of Technology) and Sertac Karaman (Massachusetts Institute of Technology) for their talks at 2021 IEEE ICRA 1st workshop "Opportunitites and Challenges with Autonomous Racing"\footnotemark[8] which contributed to the creation of Section \ref{sec_openquestios} in this survey paper.

\footnotetext[8]{https://linklab-uva.github.io/icra-autonomous-racing}

\ifCLASSOPTIONcaptionsoff
  \newpage
\fi

\newpage
\appendix
\begin{table}[h]
\centering
\caption{List of Abbreviations}
\begin{tabular}{|l|l|}
\hline
\textbf{Abbreviation} & \textbf{Definition}                     \\ \hline
A3C         & Advantage Actor-Critic                            \\ \hline
AMCL        & Adaptive Monte Carlo Localization                 \\ \hline
%AutOps      & Autonomous System Operations                      \\ \hline
AWD         & All Wheel Drive                                   \\ \hline
BVSC        & Backstepping Variable Structure Control           \\ \hline
BO          & Bayesian Optimization                             \\ \hline
CI          & Continuous Integration                           \\ \hline
CMA         & Covariance Matrix Adapation                       \\ \hline
CMA-ES      & Covariance Matrix Adapation Evolutionary Strategy \\ \hline
CNN         & Convolutional Neural Network                      \\ \hline
CPU         & Central processing unit                           \\ \hline
COP         & Center of Percussion                              \\ \hline
DDPG        & Deep Deterministic Policy Gradient                \\ \hline
DMD-MPC     & Dynamic Mirror Descent MPC                        \\ \hline
DNN         & Deep Neural Network                               \\ \hline
DP          & Dynamic Programming                               \\ \hline
EA          & Evolutionary Algorithm                            \\ \hline
ECU         & Electrical Control Unit                           \\ \hline
EHF         & H$\infty$ Filter                                  \\ \hline
EKF         & Extended Kalman Filter                            \\ \hline
ES          & Evolutionary Strategy                             \\ \hline
FSD         & Formula Student Driverless                        \\ \hline
G-G         & Maximal Lateral and Longitudinal Accelerations    \\ \hline
GA          & Genetic Algorithm                                 \\ \hline
GP          & Gaussian Process                                  \\ \hline
GPS         & Global Positioning System                         \\ \hline
GPU         & Graphical Processing Unit                         \\ \hline
HD          & High Definition                                   \\ \hline
IAC         & Indy Autonomous Challenge                         \\ \hline
ILC         & Iterativ Learning Control                         \\ \hline
IMU         & Inertial Measurement Unit                         \\ \hline
KF          & Kalman Filter                                     \\ \hline
L2R         & Learn-to-Race                                     \\ \hline
LMP         & Le Mans Prototype                                 \\ \hline
LMPC        & Learning Model Predictive Control                 \\ \hline
LSTM        & Long Short Term Memory                            \\ \hline
LPV         & Linear Parameter-Varying                          \\ \hline
MIQP        & Mixed Integer Quadratic Programming               \\ \hline
MPC         & Model Predictive Control                          \\ \hline
MPPI        & Model Predictive Path Integral Control            \\ \hline
NDT         & Normal Distribution Transform                     \\ \hline
NMPC        & Nonlinear Model Predictive Control                \\ \hline
NPC         & Non Playable Character                            \\ \hline
OCP         & Optimal Control Problem                           \\ \hline
PID         & Proportional Integral Derivative Controller       \\ \hline
POMDP       & Partially Observable Markov Decision Process       \\ \hline
QCQP        & Convex Quadratically Constrained Quadratic Problem\\ \hline
RC          & Remote Controlled                                 \\ \hline
ROS         & Robot Operating System                            \\ \hline
RL          & Reinforcement Learning                            \\ \hline
RRT         & Rapidly-exploring random tree                     \\ \hline
RTK         & Real-Time Kinematic                               \\ \hline
RNN         & Recurrent Neural Network                          \\ \hline
RWD         & Rear Wheel Drive                                  \\ \hline
SAC         & Soft Actor Critic                                 \\ \hline
SQP         & Sequential Quadratic Programming                  \\ \hline
SRMPC       & sparse Randomized MPC                             \\ \hline
SSD         & Single Shot Detection                             \\ \hline
UKF         & Unscented Kalman Filter                            \\ \hline
TMPC        & Tube MPC                                          \\ \hline
SLAM        & Simultaneous Localization and Mapping             \\ \hline
YOLO        & You Only Look Once (DNN)                          \\ \hline

\end{tabular}
\end{table}

\newpage



%\bibliographystyle{IEEEtran}

\bibliographystyle{IEEEtran}
\bibliography{Transactions-Bibliography/main.bib}



\vspace{-1cm}

\begin{IEEEbiography}[{\includegraphics[width=1in,height=1.25in,clip,keepaspectratio]{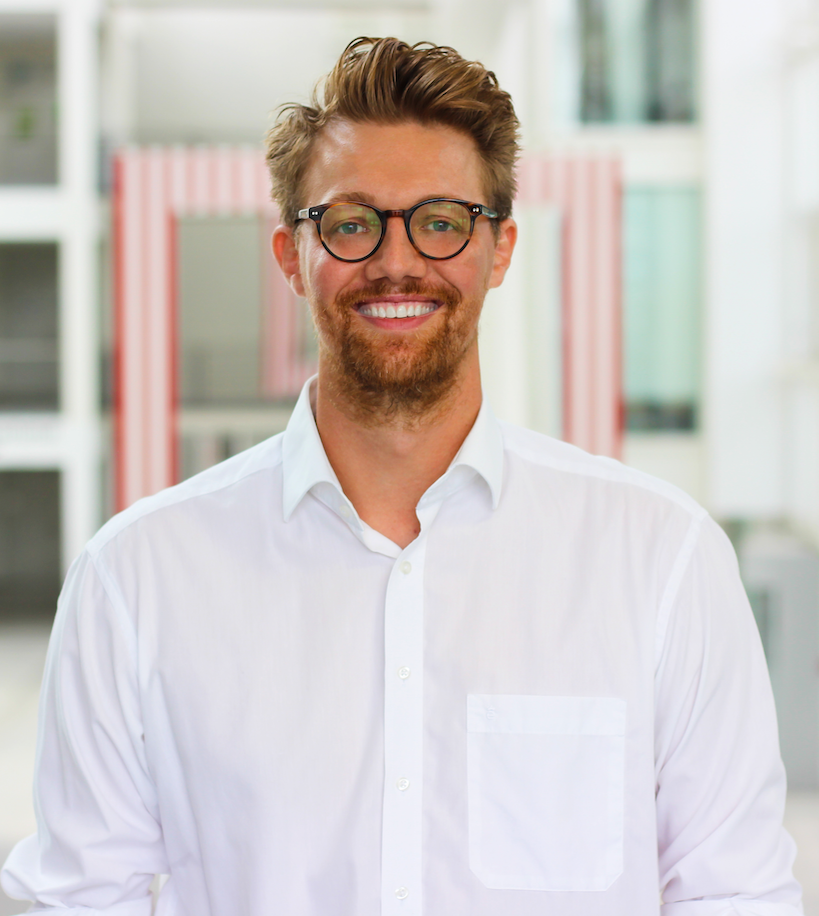}}]{Johannes Betz} earned both a B. Eng. (2011) and a M. Sc. (2012) in the field of Automotive Engineering. After he did is PhD at the Technical University of Munich (TUM) he was Postdoctoral Researcher at the Institute of Automotive Technology at TUM where he founded the TUM Autonomous Motorsport Team. He is now a postdoctoral researcher at the University of Pennsylvania where he is working at the xLab for Safe Autonomous Systems. His research is focusing on a holistic software development for autonomous systems with extreme motions at the dynamic limits in extreme and unknown environments. By using modern algorithms from the field of artificial intelligence he is trying to develop new and advanced methods and intelligent algorithms. Based on his additional studies in philosophy he extends current path and behavior planners for autonomous systems with ethical theories
\end{IEEEbiography}
\vspace{-1.5cm}

\begin{IEEEbiography}[{\includegraphics[width=1in,height=1.25in,clip,keepaspectratio]{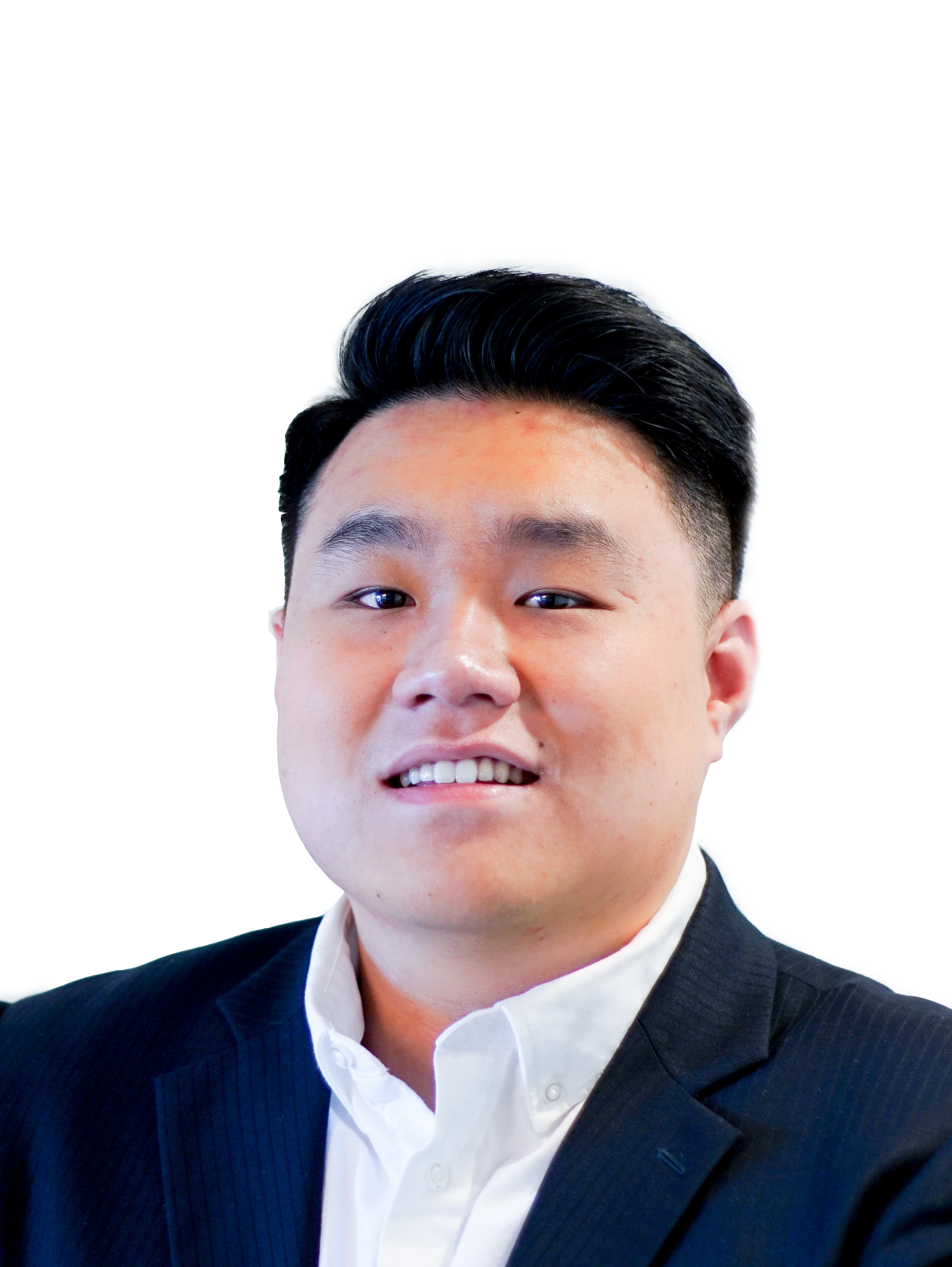}}]{Hongrui Zheng} received his B.S. degrees in Mechanical Engineering and Computer Science from Georgia Institute of Technology, and his M.S. degree in Robotics from University of Pennsylvania. He is currently a Ph.D. candidate at the University of Pennsylvania where he is working at the xLab for Safe Autonomous Systems. His research focuses on building the tools and theoretical foundations necessary to scale design, testing, and optimization of safe-autonomous systems.
\end{IEEEbiography}

\vspace{-1cm}

\begin{IEEEbiography}[{\includegraphics[width=1in,height=1.25in,clip,keepaspectratio]{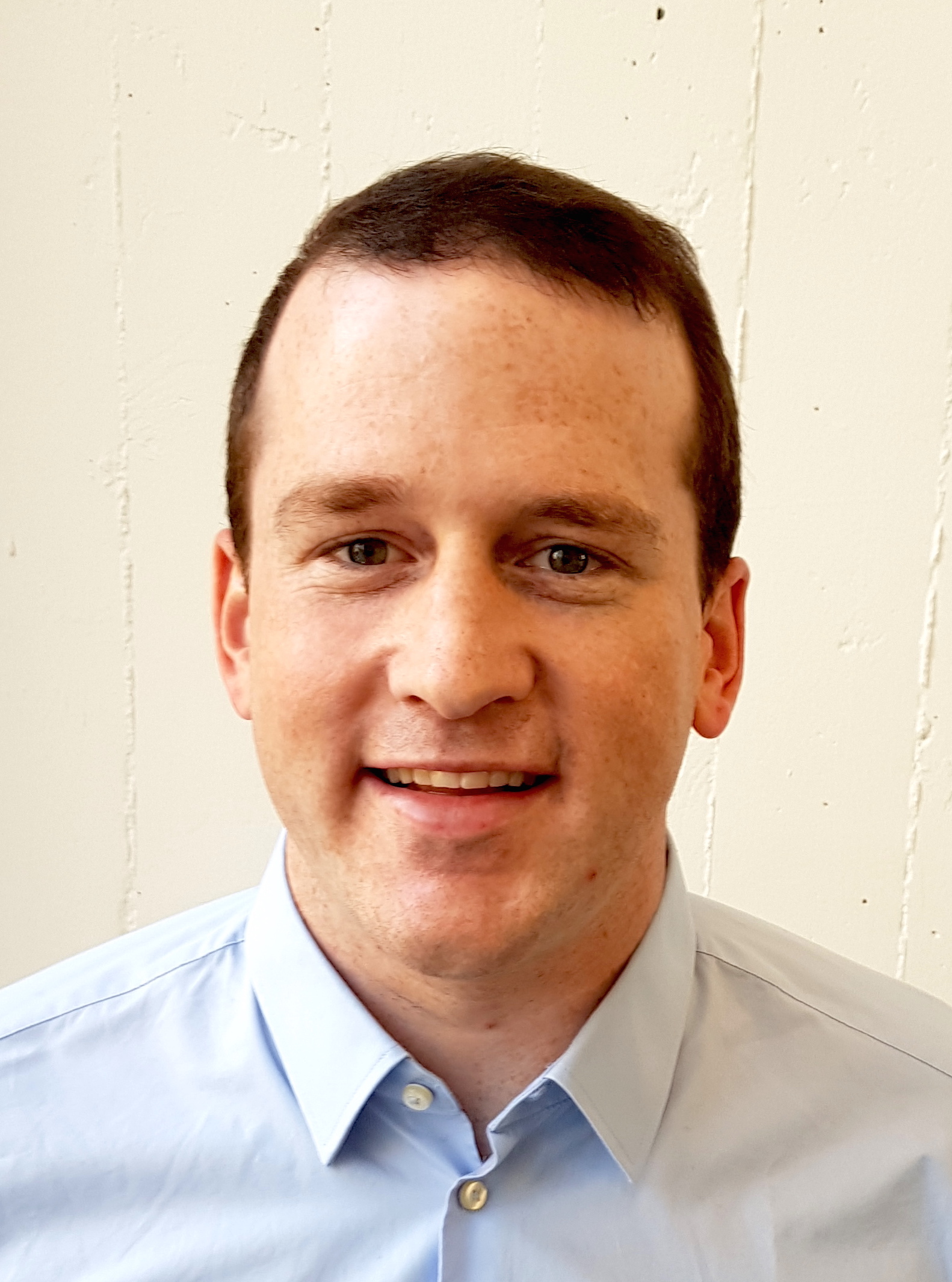}}]{Alexander Liniger} received the B.Sc. and M.Sc. degrees in mechanical engineering from the Department of Mechanical and Process Engineering, ETH Zurich, Switzerland, in 2010 and 2013, respectively, and in May 2018 successfully defended his PhD degree at the Automatic Control Laboratory, ETH Zurich, Switzerland. Currently he is a Postdoctoral Researcher in the Computer Vision Lab, ETH Zurich, Switzerland, where he is part of Luc van Gool's group working in the Toyta TRACE project. During his PhD his main research interests include model predictive control, viability theory as well as game theory and their application to autonomous driving and racing. Currently, he is investigating how control theory and computer vision can be combined to achieve end-to-end learning approaches with formal guarantees.
\end{IEEEbiography}

\vspace{-1.0cm}

\begin{IEEEbiography}[{\includegraphics[width=1in,height=1.25in,clip,keepaspectratio]{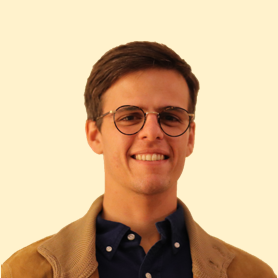}}]{Ugo Rosolia} is a postdoctoral scholar at Caltech, working with Prof. Aaron Ames and Prof. Yisong Yue. His current research focuses on high-level planning in partially observable environments and on designing control algorithms that allow autonomous systems to perform highly dynamical maneuvers while guaranteeing safety. During his PhD at the University of California Berkeley, he worked with Prof. Francesco Borrelli in the MPC lab. He developed the Learning Model Predictive Control (LMPC) strategy, which is a model-based policy iteration strategy. This strategy was used to teach an autonomous vehicle how to race.
\end{IEEEbiography}

\begin{IEEEbiography}[{\includegraphics[width=1in,height=1.25in,clip,keepaspectratio]{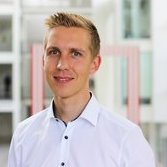}}]{Phillip Karle} is currently a PhD student at the Technical University of Munich. He received both his B.Sc. degree in 2017 and his M.Sc. degree in 2019 from the Technical University of Munich (TUM), Munich, Germany, where he is currently working towards his Ph.D. in mechanical engineering at the Institute of Automotive Technology. His main research interests include data-mining, scenario understanding, motion prediction and related applications for autonomous driving with the focus on unstructured environments.
\end{IEEEbiography}

\vspace{-2cm}

\begin{IEEEbiography}[{\includegraphics[width=1in,height=1.25in,clip,keepaspectratio]{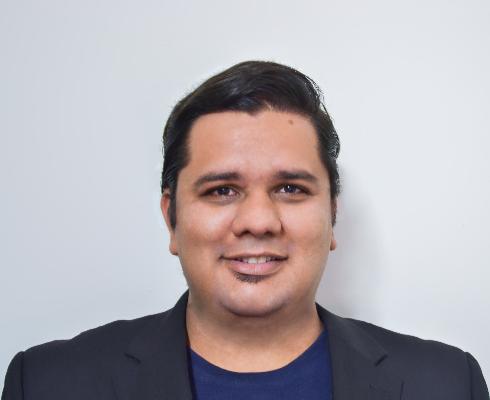}}]{Madhur Behl} is an assistant professor in the departments of Computer Science, and Systems and Information Engineering, and a member of the Cyber-Physical Systems Link Lab at the University of Virginia.  He conducts research at the confluence of Machine Learning, Predictive Control, and Artificial Intelligence with applications in Cyber-Physical Systems, Autonomous Systems, Robotics, and Smart Cities. He received his Ph.D. (2015) and M.S. (2012), in Electrical and Systems Engineering, both from the University of Pennsylvania.
\end{IEEEbiography}

\vspace{-1cm}

\begin{IEEEbiography}[{\includegraphics[width=1in,height=1.25in,clip,keepaspectratio]{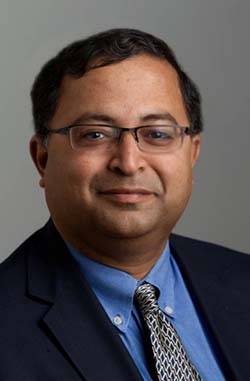}}]{Venkat Krovi} is the Michelin Endowed Chair of Vehicle Automation in the Departments of Automotive Engineering and Mechanical Engineering at Clemson University. He received his Ph.D. degree in Mechanical Engineering and Applied Mechanics from the University of Pennsylvania in 1998. 
The underlying theme of his research activities has been to take advantage of the “power of the many” (both autonomous-agents and humans) to extend the reach of human users in the dull, dirty, and dangerous environments. These efforts at analyzing and realizing the potential of distributed-autonomy and human-robot synergy has unlocked new opportunities in wide ranging applications in plant-automation, consumer electronics, automobile, defense and healthcare/surgical simulation arenas.  
\end{IEEEbiography}

\vspace{-1cm}

\begin{IEEEbiography}[{\includegraphics[width=1in,height=1.25in,clip,keepaspectratio]{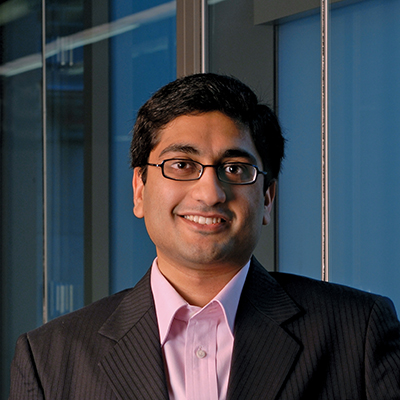}}]{Rahul Mangharam} received his Ph.D. in Electrical and Computer Engineering from Carnegie Mellon University where he also received his MS and BS.  He is an Associate Professor in the Department of Electrical and Systems Engineering at the University of Pennsylvania. He is a founding member of the PRECISE Center and directs the xLab for Safe Autonomous Systems Lab at Penn. His research is at the intersection of formal methods,  machine learning and control for medical devices,  energy efficient buildings, and autonomous systems.
\end{IEEEbiography}



\end{document}